# Accelerating Reinforcement Learning by Composing Solutions of Automatically Identified Subtasks

**Chris Drummond**                                        CDRUMMON@SITE.UOTTAWA.CA
*School of Information Technology and Engineering*
*University of Ottawa, Ontario, Canada, K1N 6N5*

## Abstract

This paper discusses a system that accelerates reinforcement learning by using transfer from related tasks. Without such transfer, even if two tasks are very similar at some abstract level, an extensive re-learning effort is required. The system achieves much of its power by transferring parts of previously learned solutions rather than a single complete solution. The system exploits strong features in the multi-dimensional function produced by reinforcement learning in solving a particular task. These features are stable and easy to recognize early in the learning process. They generate a partitioning of the state space and thus the function. The partition is represented as a graph. This is used to index and compose functions stored in a case base to form a close approximation to the solution of the new task. Experiments demonstrate that function composition often produces more than an order of magnitude increase in learning rate compared to a basic reinforcement learning algorithm.

## 1. Introduction

A standard reinforcement learning algorithm, applied to a series of related tasks, could learn each new task independently. It only requires knowledge of its present state and infrequent numerical rewards to learn the actions necessary to bring a system to some desired goal state. But this very paucity of knowledge results in a slow learning rate. This paper shows how to exploit the results of prior learning to speed up the process while maintaining the robustness of the general learning method.

The system proposed here achieves much of its power by transferring parts of previously learned solutions, rather than a single complete solution. The solution pieces represent knowledge about how to solve certain subtasks. We might call them macro-actions (Precup, Sutton, & Singh, 1997), with the obvious allusion to macro-operators commonly found in Artificial Intelligence systems. The main contribution of this work is in providing a way of automatically identifying these macro-actions and mapping them to new tasks.

This work uses syntactic methods of composition much like in symbolic planning, but the novelty arises in that the parts being composed are multi-dimensional real-valued functions. These functions are learned using reinforcement learning as part of more complex functions associated with compound tasks. The efficacy of this approach is due to the composition occurring at a sufficiently abstract level, where much of the uncertainty has been removed. Each function acts much like a funnel operator (Christiansen, 1992), so although individual actions may be highly uncertain, the overall result is largely predictable.

The subtasks are identified on the basis of strong features in the multi-dimensional function that arise during reinforcement learning. The features are not "in the world", but in the system's interaction with the world. Here, "strong" means that the features are stable (i.e. relatively insensitive to variations in the low level learning process) and easy to recognize and locate accurately early in the learning process. One important aspect of these features is that they largely dictate the shape of the function. If the features differ by a small amount, one would expect the function to differ by a small amount.

The features generate a partitioning of the function. A popular technique in object recognition, the snake (Kass, Witkin, & Terzopoulus, 1987; Suetens, Fua, & Hanson, 1992), is used to produce this partition. In object recognition, the snake produces a closed curve that lies along the boundary of an object, as defined by edges in an image. In this application, the snake groups together sets of features to define a region of the function. The boundary of the region is a low order polygon, demarcating an individual subtask. This is repeated until the whole function is covered. The polygons are converted into discrete graphs, a vertex of the polygon becoming a node of the graph. Merging these graphs produces a composite graph representing the whole task.

The composite graph is used to control the transfer by accessing a case base of previously learned functions. The case base is indexed by graphs. The relevant function is determined by matching a subgraph of the composite graph with one acting as an index to a case. The associated functions are transformed and composed to form a solution to the new task. This is used to reinitialize the lower level learning process. It is not necessary for the transfer to produce an exact solution for the new task. It is sufficient that the solution is close enough to the final solution often enough to produce an average speed up. Reinforcement learning will further refine the function and quickly remove any error.

This paper demonstrates the applicability of transfer in two different situations. In the first, the system learns a task for a particular goal position and then the goal is moved. Although the function itself will change significantly, the partition generated on the initial task can be used to compose the function for the new task. In the second situation considered, the system is placed in a different environment within the same domain. Here, a new partition has to be extracted to control the composition process.

This paper unifies and significantly extends previous work by the author (Drummond, 1997, 1998). Additional work has largely focussed on removing some of the limitations inherent in the partitioning approach introduced in Drummond (1998). One limitation of the original approach was that the snake could only extract polygons that were rectangles. This paper relaxes this restriction, allowing it to be applied to a different environment within the same domain and to a different task domain. Although lifting this restriction removes some desirable bias, the experiments demonstrate that none of the efficacy of the original system is lost. Further, the results are more broadly obtained on the larger set of related tasks and in a different domain. Overall, the function composition approach often produces more than an order of magnitude increase in learning rate when compared to a basic reinforcement learning algorithm.

The rest of the paper begins with Section 2 giving a very high level discussion of the approach taken. Section 3 gives a more in depth discussion of the techniques used. Sections 4 and 5 present and analyze the experimental results. Subsequent sections deal with limitations and related research.

## 2. An Overview

The intent of this section is to appeal to the intuitions of the reader, leaving much of the detail to later sections in the paper. The subsections that follow will demonstrate in turn: that there are features in the function produced by reinforcement learning; that graphs based on these features can be used to control the composition of the function pieces; that these features are easy to detect early in the learning process; that these features exist in multiple domains.

### 2.1 Features in the Reinforcement Learning Function

This overview begins with a very high level introduction to reinforcement learning and the function it produces. It will show that there are features in this function which can be extracted and converted into a graphical representation.

One of the experimental test beds used is this paper is a simulated robot environment of different configurations of interconnected rooms. The robot must learn to navigate efficiently through these rooms to reach a specified goal from any start location. Figure 1 shows one example with 5 rooms and the goal in the top right corner. The robot's actions are small steps in any of eight directions, as indicated by the arrows. Here, the location, or state, is simply the robot's $x$ and $y$ coordinates. The thin lines of Figure 1 are the walls of the rooms, the thick lines the boundary of the state space.

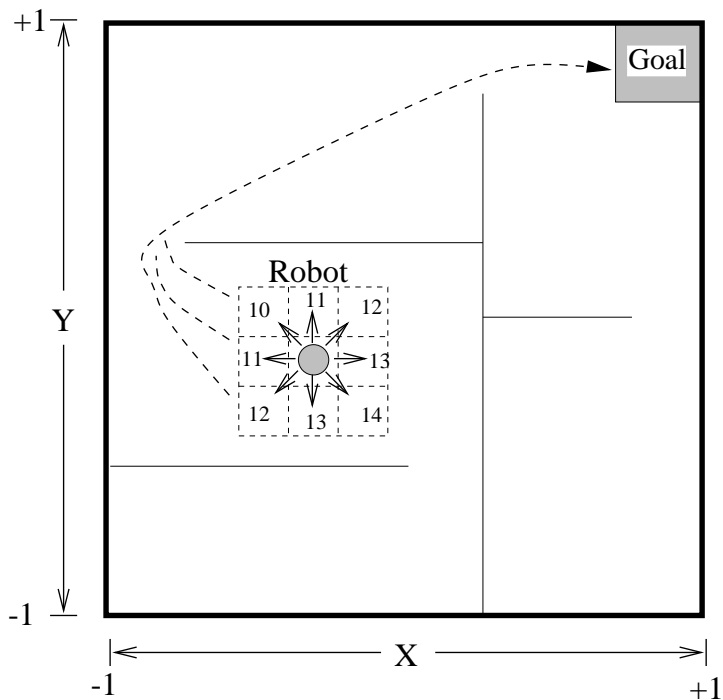

Figure 1: Robot Navigating Through a Series of Rooms

If each action is independent of preceding actions, the task becomes one of learning the best action in any state. The best overall action would be one that takes the robot immediately to the goal. But this is only possible in states close to the goal. Suppose the robot is in a particular state and that the number of steps to goal from each of its neighboring states is known, indicated by the numbered squares surrounding the robot in Figure 1. Then a one step look ahead procedure would consider each step and select the one that reaches the neighboring state with the shortest distance to goal. In Figure 1 the robot would move to the state 10 steps from goal. If this process is repeated, the robot will take the shortest path to goal. In practice we must, of course, learn such values. This can be done using some type of reinforcement learning (Watkins & Dayan, 1992; Sutton, 1990) which progressively improves estimates of the distance to goal from each state until they converge to the correct values.

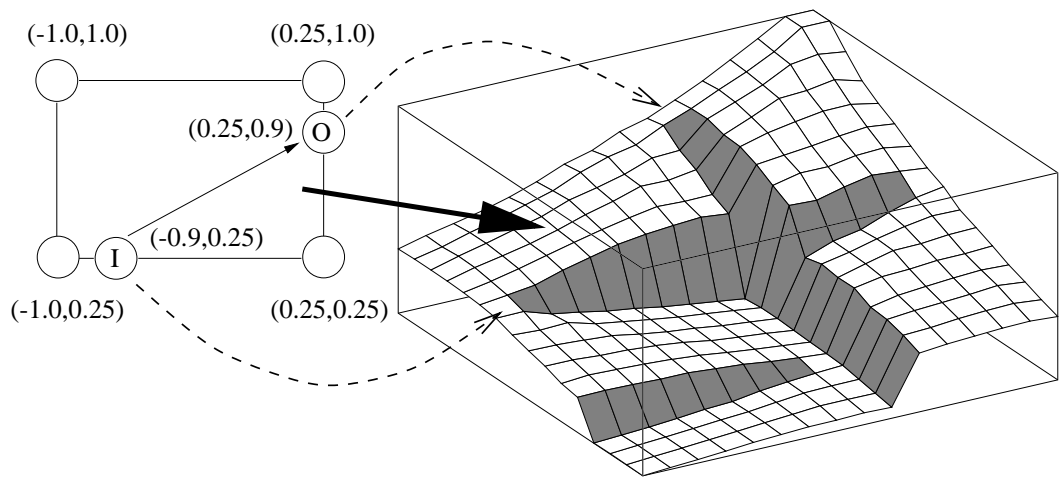

Figure 2: The Value Function Obtained Using Reinforcement Learning

The function shown in Figure 2 is called the value function. Subsequently, the term function will mean the value function unless otherwise indicated. The function is the result of reinforcement learning on the problem of Figure 1, but instead of it representing the actual distance to goal, it represents essentially an exponential decay with distance to goal. The reasons for this will be made clear in Section 3.1. The shaded areas represent large gradients in the learned function. Comparing this to the environment shown in Figure 1, it is apparent that these correspond to the walls of the various rooms. These are the *strong features* discussed in this paper. They exist because of the extra distance for the robot to travel around the wall to reach the inside of the next room on the path to the goal. These features are visually readily apparent to a human, so it seems natural to use vision processing techniques to locate them.

An edge detection technique called a snake is used to locate these features. The snake produces a polygon, in this instance a rectangle, locating the boundary of each room. The doorways to the room occur where the differential of the function, along the body of the snake, is at a local minimum. The direction of the differential with respect to edges of the polygon, associated with the walls of the room, determines if it is an entrance or an exit. A positive gradient into the room indicates an entrance; a positive gradient out of the room indicates an exit. From this information, a plane graph, labeled with an $(x, y)$ coordinate for each node, is constructed. Figure 2 shows one such example, for the room at the top left corner of the state space, subsequent graphs will not show the coordinates. Nodes corresponding to the doorways are labeled "I" or "O" for in and out respectively; their positions on the function are indicated by the dashed arrows.

## 2.2 Composing Function Pieces

This overview continues by showing how the graphs, extracted from the features in the function learned by reinforcement learning, can be used to produce a good approximation to the solution for a new goal position. The left hand side of Figure 3 shows plane graphs for all the rooms (ignore the dashed lines and circles for now). The node representing the goal is labeled "G". A directed edge is added from "I" to "O" or "I" to "G", as appropriate. Associated with this edge is a number representing the distance between the nodes. This is determined from the value of the function at the points of the doorways. Each individual graph is then merged with its neighbor to produce a graph for the whole problem, the right hand side of Figure 3. The doorway nodes have been relabeled to "D". The composite graph represents the whole function. Each individual subgraph represents a particular part of the function. This information is stored in a case base. Each subgraph is an index and the corresponding part of the function is the case.

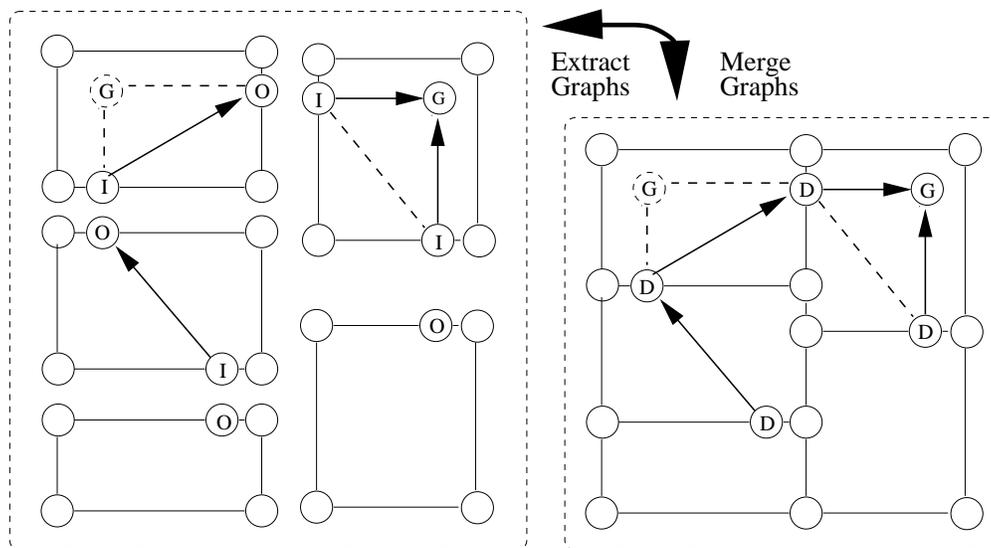

Figure 3: Graphical Representation

Now suppose the goal is moved from the top right corner to the top left corner of the state space. Reinforcement learning in its most basic form would be required to learn the new function from scratch. In this work if the goal is moved, once the new goal position is known, the node representing the goal can be relocated. The new goal position is shown as the dashed circle in Figure 3. The edges connecting the doorways and the goal are changed to account for the new goal position. The dashed lines representing the new edges replace the arrows in the same subgraph. To produce a new function, the idea is to regress backwards from the goal along these edges. For each edge, the small subgraph containing the edge is extracted. The extracted subgraph is used to index the case base of functions. The retrieved function is transformed and added to the appropriate region of the state space to form the new function.

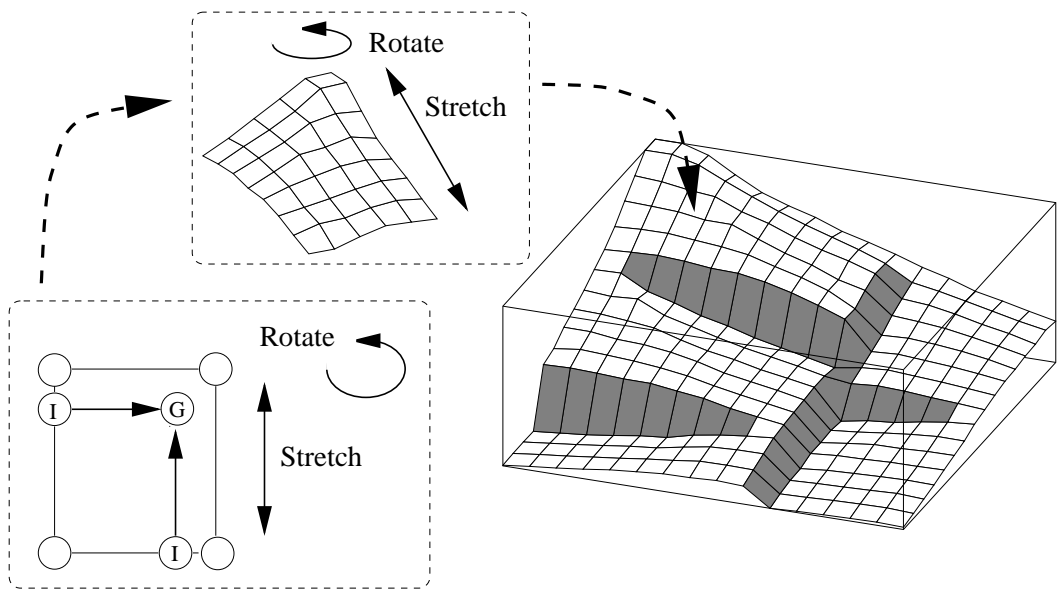

Figure 4: Function Composition

In this example, some of the existing subgraphs match the new configuration. The two that do not are the subgraph originally containing the goal and the subgraph now containing the goal. It is certainly possible to exchange these two, using an appropriate transform. But other graphs in the case base may better match the new task. The best match for the subgraph containing the new goal is, in fact, the subgraph for the goal in the original problem. To fit this to the new task, the plane graph is rotated and stretched slightly in the new $x$ direction by changing the coordinates of its nodes, see Figure 4. Then this same transformation is applied to the function. But for the room containing the original goal, a case obtained when solving another task is a better match. The other three rooms use the functions from the original problem, since changing the goal position has little effect on the actions taken. In fact, only the height of the functions must be changed. This is simply a multiplication by a value representing the distance to goal from the "O" doorway (this is discussed in detail at the end of Section 3.3). Because the matching of the subgraphs allows some error and asymmetric scaling may be used, the resulting function may not be exact. But as the experiments will demonstrate, the function is often very close and further reinforcement learning will quickly correct any error.

The new position of the goal must be established before the graph can be modified and function composition can occur. The system is not told that the goal has moved, rather it discovers this by determining that it is no longer at the maximum of the existing function. There is some uncertainty in the exact boundary of the original goal. The robot may reach a state which it believes is part of the original goal region, but fail to detect it even if the goal has not moved. To be reasonably certain that the goal has in fact moved, this is required to occur ten times with no intervening occurrence of the goal being detected at the maximum.

The system then composes a search function, by assuming a particular room contains the goal. Search functions are also produced by composing previously learned functions. However, for the room assumed to contain the goal the function is a constant. This does not bias the search to any particular part of the room and allows some limited learning to encourage exploration of the room. The search function drives the robot into the room from anywhere else in the state space. If it fails to find the goal after a fixed number of steps, a new search function is composed with another room assumed to contain the goal. This process is repeated until the goal has been located ten times, this ensures a good estimate of the "center of mass" of the goal. The "center of mass" is used as the new position of the goal node in the composite graph. Requiring that the old goal or new goal positions are sampled a fixed number of times has proven to be effective in the domains discussed in this paper. Nevertheless, it is a somewhat ad hoc procedure and will be addressed in future work, discussed in Section 6.2.

## 2.3 Detecting Features Early

In the previous section, the existing task and the new task were strongly related, the walls and doorways were fixed and only the goal position was different. In this section, no such relationship is assumed. The robot is faced with a brand new task and must determine what, if any, relationship exists between the new task and any previous tasks.

The experimental testbed is again a simulated robot environment, but this time the problem is simplified to just an inner rectangular room and an outer L-shaped room. Figures 5 and 6 show two possible room configurations. Again, the thin lines are the walls of the room, the thick lines the boundary of the state space. Suppose the robot had already learned a function for the "Old Task" of Figure 5. We would hope that we could adapt the old solution to fit the closely related "New task" of Figure 6.

The steps, in this example, are essentially those in the previous one. But now as the learning process is started afresh, there are no features and the system must wait until they emerge through the normal reinforcement learning process. Then we can proceed much as before. First a graph for the inner room is extracted. The best matching graph in the case base from the old task is rotated and stretched to fit the new task. Next a matching graph for the outer L-shaped room is rotated and stretched around the larger inner room. The same transforms are then applied to the associated functions, any height adjustments carried out and the functions composed to form an approximate solution to the new task.

In this example, the first step in the process is to locate the goal. As there is no partition to aid the search, the initial value function is set to a mid-range constant value (see Figure 7). This allows some limited learning which encourages the system to move away from regions it has explored previously, to prevent a completely random walk through state space. Once the goal is located, the learning algorithm is reinitialized with a function for the same goal position but no walls (see Figure 8). If such a function does not exist in the case base, any rough approximation could be used instead. The "no walls" function is not used exactly as stored in the case base. The difference between the goal and the rest of the state space is reduced by scaling the function then adding a constant. This reduces the "bias" of the function, allowing the learning algorithm to alter it relatively easily as new information becomes available.

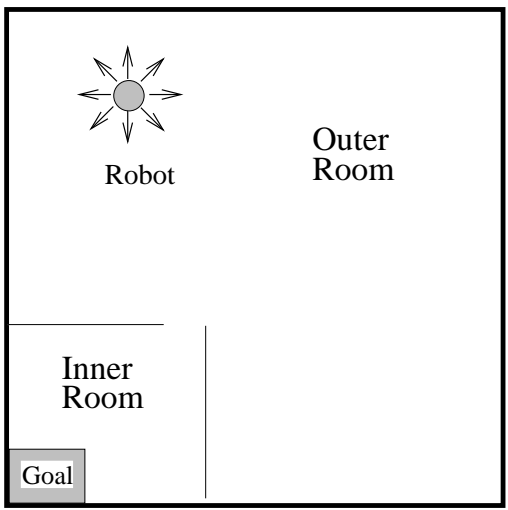

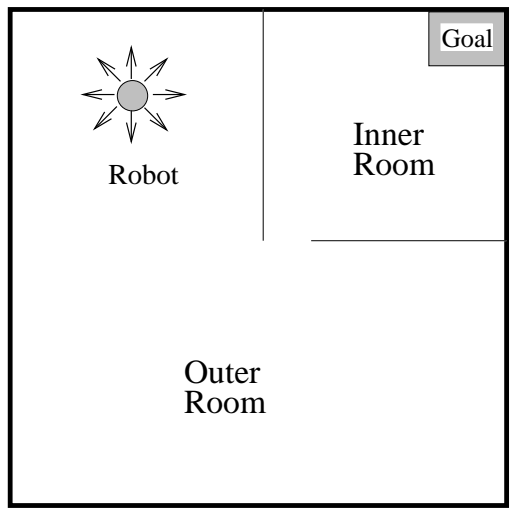

Figure 5: The Old Task                    Figure 6: The New Task

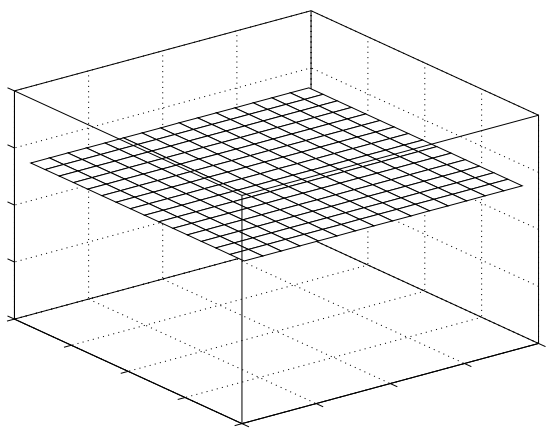

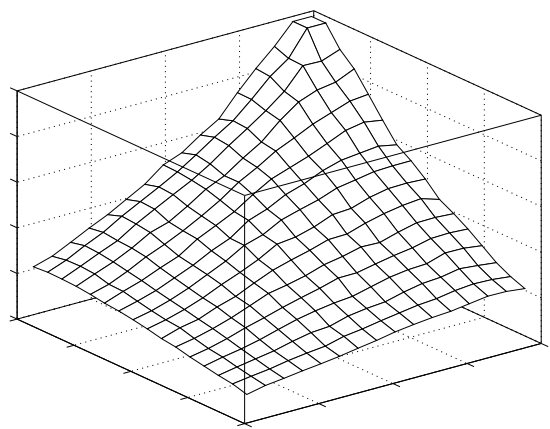

Figure 7: Start Function                  Figure 8: Intermediate Function

Figure 9 shows the resultant function about 3000 exploratory steps from the beginning of the learning process. Again, the large gradients associated with the walls are readily apparent. Figure 10 shows the function for the new task if it had been allowed to converge to a good solution. Both functions have roughly the same form, the large gradients are in the same position, although learning the latter took some 200,000 steps. After the "no walls" function is introduced the features take some time to clearly emerge. The snake will typically filter out features that are too small and not well formed. Additional filtering at the graphical level further constrains acceptable features. The total set of features must produce a consistent composite graph, the doorways from different subgraphs must align and the graph must overlay the complete state space. There must also be a matching case in the case base for every subtask. Many of these checks and balances will be removed when the iterative updating technique of Section 6.2 is incorporated.

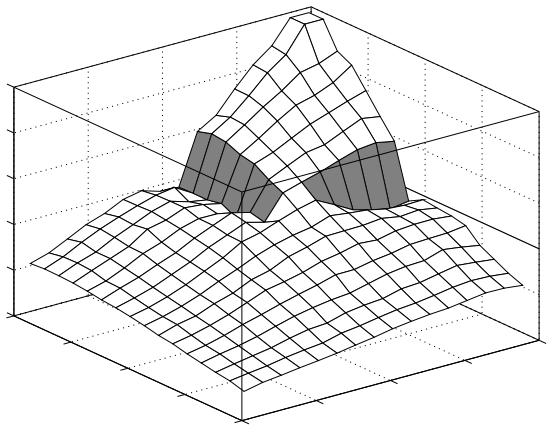

Figure 9: Early Function

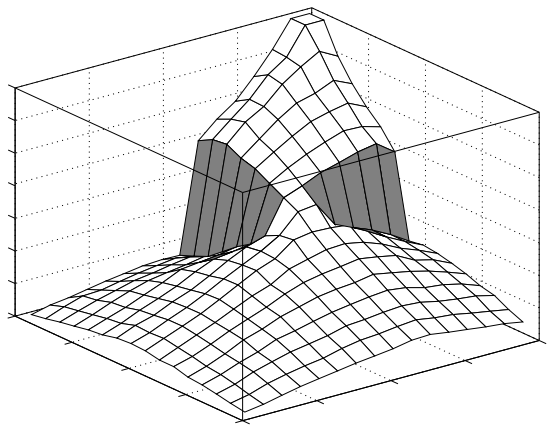

Figure 10: New Task Function

## 2.4 A Different Task Domain

The previous sections dealt with a simple robot navigation problem. This section demonstrates that these features also exist in a quite different domain, that of a two degrees of freedom robot arm, as shown in Figure 11. The shoulder joint can achieve any angle between $\pm\pi$ radians, the elbow joint any angle between $\pm\pi/2$ radians, zero is indicated by the arrows. If the arm is straight and the shoulder joint rotated, the elbow joint will describe the inner dotted circle, the hand the outer dotted circle. There are eight actions, small rotations either clockwise or anti-clockwise for each joint separately or together. The aim is to learn to move the arm efficiently from any initial position until the hand reaches the goal on the perimeter of the arm's work space.

The state space, for the purposes of reinforcement learning, is the configuration space for the arm, sometimes called the joint space (see Figure 12). The $x$-axis is the angle of the shoulder joint, the $y$-axis the elbow joint. The eight actions when mapped to actions in the configuration space become much like the actions in the robot navigation problem, as shown by the shaded diamond (labeled Arm) in Figure 12. To map an obstacle in the work space to the configuration space, one must find all pairs of shoulder and elbow angles blocked by the obstacle. The obstacles in this space become elongated to form barriers much like the walls in the experiments of the previous sections. If this is not clear, imagine straightening the arm in the work space and rotating it such that it intersects one of the obstacles, the middle dotted line in Figure 11. The arm can then be rotated at the shoulder joint with a roughly linearly proportional rotation in the elbow joint, but in the opposite direction, such as to keep it intersecting the obstacle. This produces the "wall" in the configuration space. This linearity holds only for small objects not too far from the perimeter of the work space. More complex, larger objects, would result in more complex shapes in the configuration space. At the moment the feature extraction method is limited to these simpler shapes, this will be discussed further in Section 6.

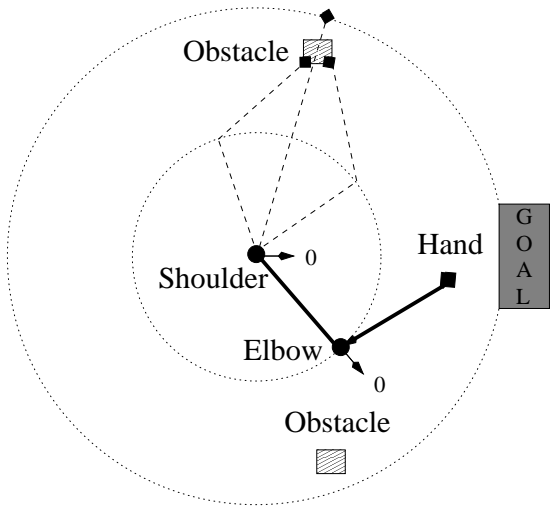

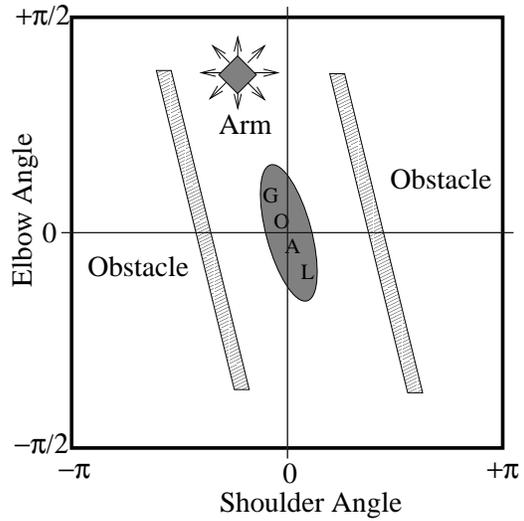

Figure 11: Work Space                 Figure 12: Configuration Space

The reinforcement learning function produced by this problem is shown in Figure 13. As before the features are shaded for clarity. The large gradient associated with the obstacle on the left hand side of the configuration space can be clearly seen. There is a similar large gradient associated with the obstacle on the right hand side of the configuration space. Again, these features can be used to control the composition of functions if the goal is moved or for a different task in the same domain.

## 3. Details of the Techniques Used

This section will discuss in more detail the techniques used. These include: reinforcement learning to produce the initial function, snakes to extract the features producing the graph, and the transformation and composition of the subgraphs, and their corresponding functions, to fit the new task.

### 3.1 Reinforcement Learning

Reinforcement learning typically works by refining its estimate of expected future reward. In goal-directed tasks, such as the ones investigated here, this is equivalent to progressively improving the estimate of the distance to goal from each state. This estimate is updated by the best local action, i.e. the one moving the robot, or arm, to the new state with the smallest estimated distance. Early in the learning process, only states close to the goal are likely to have accurate estimates of true distance. Each time an action is taken, the estimate of distance at the new state is used to update the estimate at the old state. Eventually this process will propagate back accurate estimates from the goal to all other states.

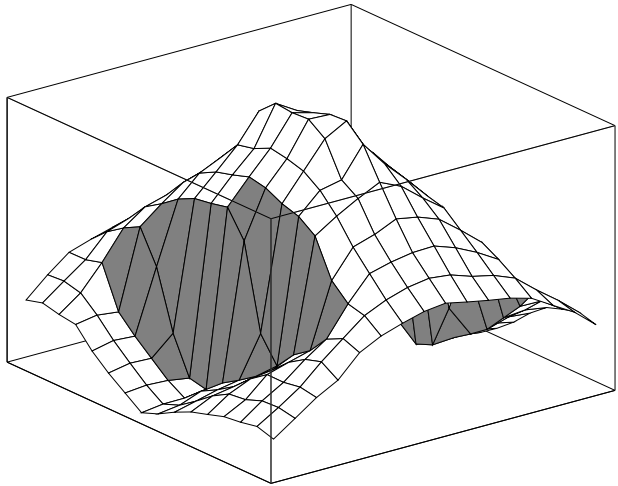

Figure 13: The Robot Arm Function

Rather than directly estimating the distance to goal, the system uses the expected discounted reward for each state $E[\sum_{t=1}^{\infty} \gamma^t r_t]$. The influence of rewards, $r_t$, are reduced progressively the farther into the future they occur by using a $\gamma$ less than one. In this work, the only reward is for reaching the goal. So the farther the state is from the goal the smaller the value. The use of an expectation allows the actions to be stochastic, so when the robot, or arm, takes a particular action in a particular state, the next state is not always the same.

To carry out reinforcement learning, this research uses the Q-learning algorithm (Watkins & Dayan, 1992). This algorithm assumes the world is a discrete Markov process, thus both states and actions are discrete. For each action $a$ in each state $s$, Q-learning maintains a rolling average of the immediate reward $r$ plus the maximum value of any action $a'$ in the next state $s'$ (see Equation 1). The action selected in each state is usually the one with the highest score. But to encourage exploration of the state space, this paper uses an $\epsilon$-greedy policy (Sutton, 1996) which chooses a random action a fraction $\epsilon$ of the time. The only effect that function composition has on the Q-learning algorithm is that the initial value for each state-action pair is set to some value other than zero.

$$Q_{s,a}^{t+1} = (1-\alpha)Q_{s,a}^t + \alpha(r + \gamma max_{a'} Q_{s',a'}^t) \tag{1}$$

The Q-function over state and action is usually referred to as the action-value function. In this paper, it is the action-value function that is transformed and composed to form a solution to the new task. The value function, discussed in previous sections and shown in the figures, is the maximum value of the Q-function. It is used to generate the partition and associated graphs needed to control the process.

Watkins and Dayan (1992) proved that Q-learning will converge to the optimal value with certain constraints on the reward and the learning rate $\alpha$. The optimal solution is produced by taking the action with the greatest value in any state. So, for goal-directed tasks, a greedy algorithm will take the shortest path to the goal, once learning is complete. The extension to continuous spaces may be done using function approximation. The simplest method, and the one used here, is to divide the state dimensions into intervals. The resulting action-value function has cells representing the average Q-value of taking each action from somewhere within a region of the state space. In off-line learning, where any action in any state can be executed, this representation has been proven to converge (Gordon, 1995). In on-line learning, where the current state is determined by the environment, this approach is generally successful, but there exists no proof of its convergence.

## 3.2 Feature Extraction

Feature extraction uses a vision processing technique that fits a deformable model called a snake (Kass et al., 1987) to edges in an image. After initializing the snake, the process iterates until external forces, due to the edges, balance internal forces in the snake that promote a smooth shape. Here, the external forces are due to steep gradients in the value function. As a piecewise constant function approximator is used, a smoothed cubic b-spline is fitted to the value-function and used to generate the necessary derivatives. The left hand side of Figure 14 is the gradient of the value function shown in Figure 9 when extracting features early in the learning process. The system has added a gradient around the border to represent the state space boundary.

To locate the features, a curve is found that lies along the ridge of the hills, a local maximum in the differential. On the right hand side of Figure 14, the dashed lines are contour lines for the small inner room as indicated. The bold lines, on the right hand side of Figure 14, are the snake at different stages of the process. The snake is first positioned approximately in the center of the room, the innermost circle. It is then expanded until it abuts on the base of the hills. Now to simplify the exposition, we can imagine that the snake consists of a number of individual hill climbers spread out along the line representing the snake, indicated by the small white circles. But instead of being allowed to climb independently, their movement relative to each other is constrained to maintain a smooth shape. When the snake reaches the top of the ridge, it is further constrained to be polygon – in this instance a quadrilateral – the outside dark line in Figure 14. At this point, it will tend to oscillate around an equilibrium position. By limiting the step size the process can be brought into a stationary state. A more detailed mathematical treatment of this approach is given in Appendix A.

The polygon forms the "skeleton" for the graph, as shown at the top left of Figure 14. Nodes in a graph correspond to vertices of the polygon and to the doorways and the goal. Looking at the gradient plot, the doorways are regions with a small differential between the ridges. Their locations can be determined from the magnitude of the gradient along the boundary of the polygon. In this example, a node is added for the goal (labeled G) and this is connected to the "in" doorway (labeled I). The polygon delimits a region of the state space, and therefore a region of the action-value function. This becomes a case in the case base, and the corresponding graph its index. Constraining the snake to be a polygon is done for two reasons. Firstly, the vertices are needed to produce nodes in the plane graphs, which are important part of the matching process. Secondly, the additional constraint results in a more accurate fit to the boundaries of the subtask. This, in turn, results in a more accurate solution after function composition.

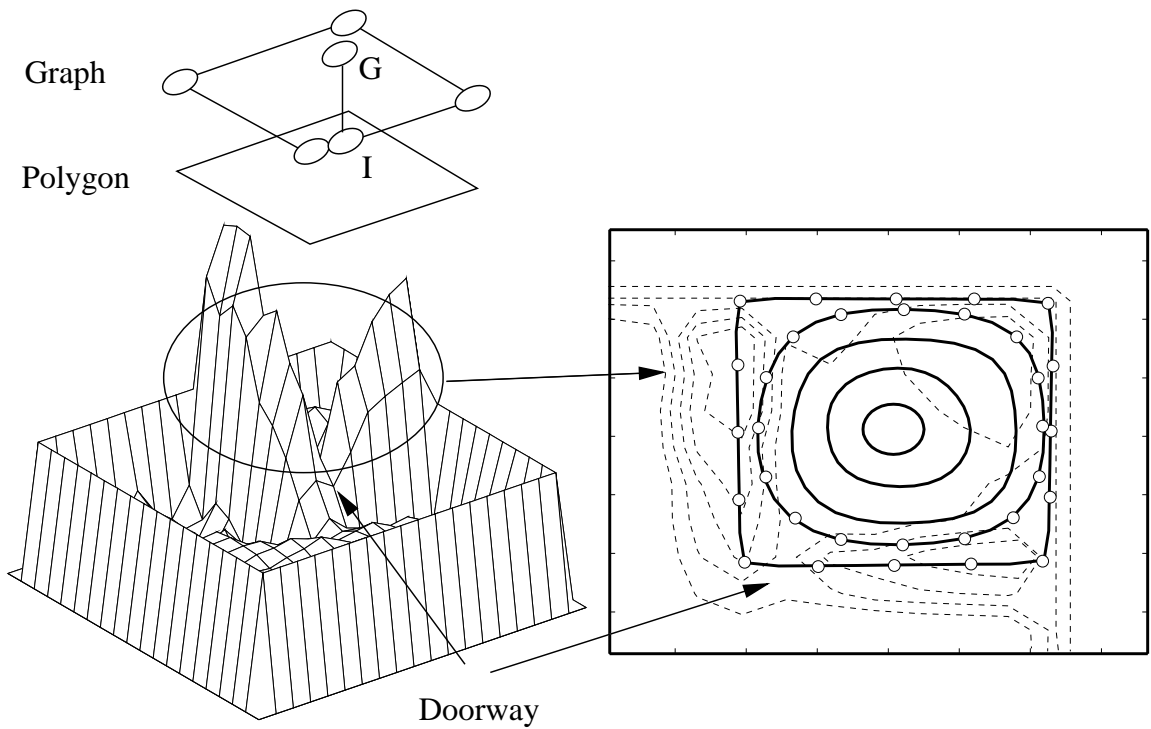

Figure 14: The Gradient and Resultant Polygon (Left) Extracted by the Snake (Right)

### 3.2.1 Three Extensions to the Snake Approach

This section introduces three extensions to the basic snake approach to facilitate the extraction of features.

The first extension affects the direction the snake moves when hill climbing the gradient. In normal hill climbing, each step is taken in the direction of steepest ascent, the step size being determined by the size of the differential. Roughly, this translates into forces at points along the body of the snake. Each force points in the direction of steepest ascent locally, but interacts with other forces through the various shape constraints. Looking at the gradient function and contour lines of Figure 14, there is a steep slope leading to the top of each ridge. But there is also a significant slope along each ridge away from the doorway towards the boundary of the state space. Thus the force on a single point on the body of the snake is not directly towards the top of the ridge but turned towards its apex, as indicated by the bold black arrow on the left hand side of Figure 15.

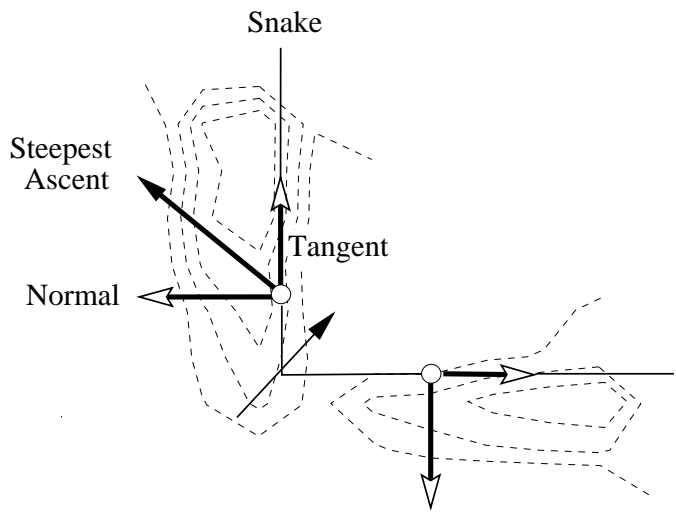

Figure 15: Controlling Forces on the Snake

This force can be broken into two components with respect to the snake, a normal and a tangential force. The latter force acts along the body of the snake. Once the shape is constrained to be a quadrilateral, this will cause the relevant side to shrink. This effect will be partially counteracted by the force towards the top of the ridge on the adjacent side of the quadrilateral. But the net result will be a shrinking of the two sides associated with the ridges inwards until the forces are balanced. This will push the corner of the quadrilateral near the doorway inwards, as indicated by the thin black arrow in Figure 15. In an extreme case, this might cause the snake to collapse into something close to a triangle. But the more likely outcome will just be a degradation of the accuracy of registration of the ridges.

Drummond (1998) prevented this degradation of the accuracy by restricting the snakes to rectangular shapes. But with the weakening of this constraint to more general polygons, this effect again becomes a problem. This problem is addressed by removing the component of the force tangential to the snake. Then hill climbing is always in the direction of the normal. This does not significantly restrict the motion of the snake: all that is being removed is the component along the body of the snake. Thus it mainly prevents the stretching and shrinking of the snake due to the gradient.

The second extension controls the way the snake is expanded to reach the base of the hills. Drummond (1998) used a ballooning force, as introduced by Cohen and Cohen (1993). But problems arose when extending the system to deal with more general shapes than rectangles, such as the outer L-shaped room in Figure 6. The ballooning force expands the snake in directions normal to its body. One deleterious effect of this is if the snake contacts a sharp external corner, such as that of the inner room, the force tends to push the snake through the corner. This can be seen in Figure 16; the bold continuous lines are the snake; the bold dashed lines are the ridges. If we imagine starting off with a circular snake in the middle of the L-shaped outer room, by the time it reaches the walls of the inner room the sides of the snake are roughly perpendicular to the ridges. Thus there is little to restrain the expansion of the snake and it passes completely through the walls of the inner room.

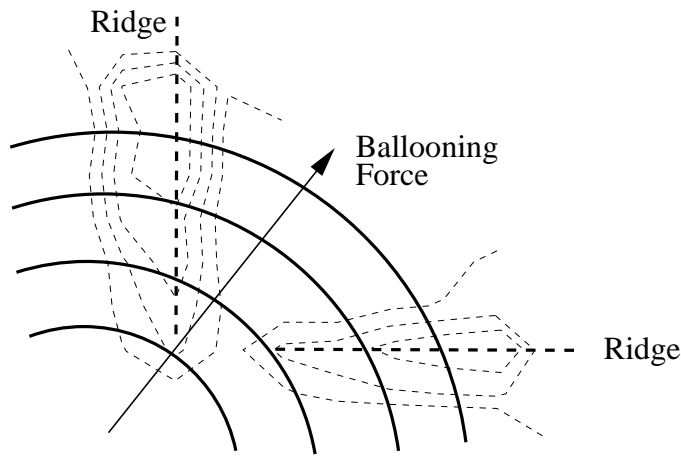

Figure 16: Using the Ballooning Force

The approach adopted here is analogous to the flow of mercury. If we imagine starting somewhere in the middle of the L-shaped room and progressively adding mercury, it would tend to fill up the lower regions of the valley first and reach the bases of the hills roughly at the same time. The analogy of mercury is used as it has a high surface tension preventing it from flowing through small gaps in the edges associated with doorways. To increase the effectiveness of this idea, the absolute value of the differential of the gradient is thresholded, values above the threshold being set to one those below to zero. It is then smoothed with a truncated Gaussian, as shown in Figure 17. Smoothing and thresholding are commonly used techniques in machine vision (Tanimoto, 1990). They are typically used to remove noise, but here the aim is to strongly blur the thresholded image. This produces bowls associated with each room. In this example, the smoothing has almost completely obscured the presence of the doorway, although this is generally not the case.

The snake is initialized as a small circle at the minimum of one of these bowls. This is shown as the circle in the middle of Figure 18, where the dashed lines are the contour lines of this function. It then flows outwards, so as to follow the contour lines of the bowl; the largest component of the flow being in the direction of the arrows in Figure 18. This is achieved by varying the force normal to the body of the snake according to the height difference with the average height of the snake. Thus points along the snake which are higher than average tend to get pushed inwards, those lower pushed outwards. The surface tension of the mercury is produced by various smoothing constraints on the first and second differentials of the snake (see Appendix A).

The third extension limits changes in the shape of the snake as it expands from its initial position to reach the base of the hills. The smoothness constraints on the snake, that give the mercury-like properties, prevent the snake flowing through the gaps associated with the doorways. But even this proved insufficient if the width of the rooms and the width of doorways were of similar sizes. In Figure 12, looking at the "room" on the left hand side of the configuration space of the robot arm, the "doorway" and the "room" at the top are of similar width. Increasing the surface tension of the mercury sufficiently to prevent flow through the doorways also prevents the flow to the top of the room.

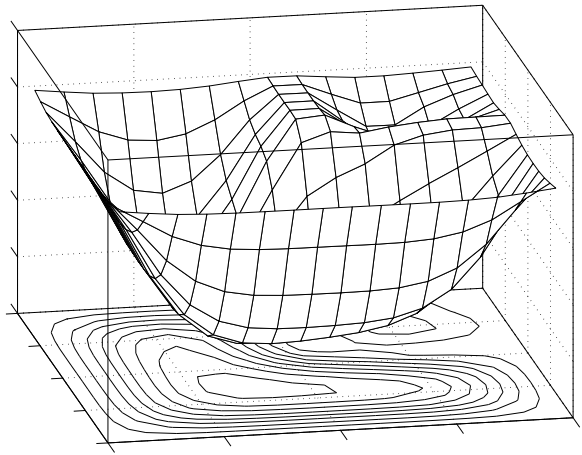

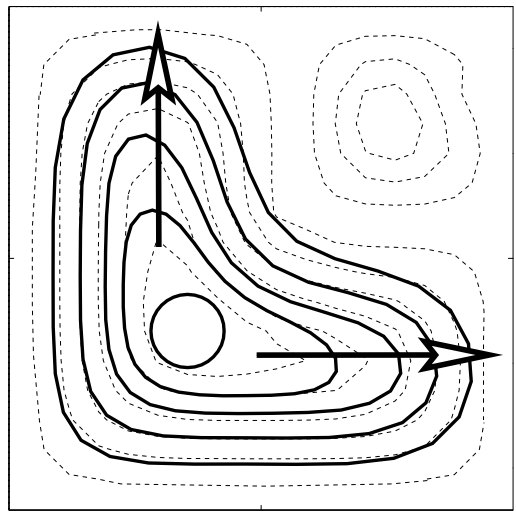

Figure 17: Smoothed Function          Figure 18: Mercury Flow

The solution is to limit the amount the snake can change its shape as it grows. This is achieved by constraining how much the second differential of the snake can change from step to step. In Figure 18, it is apparent that the snake takes up a good approximation to the shape of the room some time before it reaches the ridges. If the shape can be locked-in before reaching the ridges, the problem just described can be avoided. When the snake is initialized, the only constraint is smoothness. As the snake is expanded, this smoothness constraint is progressively weakened and the curvature constraint progressively strengthened. This progressively locks in the shape while still allowing the snake to make small local adjustments to better fit the features.

The extensions, discussed in this section, either modify the traditional forces that act on the snake or add new ones. There are also forces associated with knot spacing and drag. How the snake moves, with each iteration, depends on the vector addition of these forces. The sum acts to accelerate the body of the snake which has both mass and velocity, and therefore momentum. A schematic representation of these forces is shown in Figure 19; a more detailed mathematical description is given in Appendix A. The dashed line represents the body of the snake; the arrows are the forces applied to one point on the body. The snake is a parameterized function, given by $\hat{f}(s) = (x(s), y(s))$ where $x(s)$ and $y(s)$ are individual cubic b-splines giving the $x$ and $y$ coordinates associated with a variable $s$ along the body of the snake. The circles represent points equi-distant in $s$ but not necessarily in $x$ and $y$. These points are kept roughly the same Euclidean distance apart in $x$ and $y$ due to the knot spacing force. The momentum, although not strictly a force, encourages the point to move in constant direction; the drag opposes any motion. The stiffness encourages the snake to maintain a smooth shape. The overall stiffness is reduced as the snake grows, to keep its flexibility per unit length roughly constant, and is also controlled locally to maintain its shape.

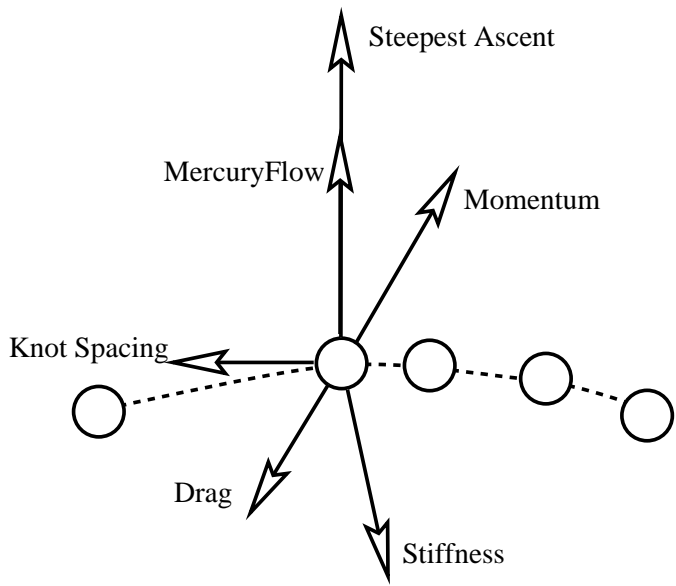

Figure 19: The Forces on the Snake

The following is an algorithmic summary of the processing of the snake:

- Initialize the coefficients to produce a circular snake in the middle of a room.

- Iterate until the forces are roughly in equilibrium and the snake oscillates around a stationary value.

- Modify the stiffness to enforce the polygonal constraints

- Iterate for a further 25 steps increasing the momentum and drag at each step to reduce the oscillation to a small value.

- Use the final position of the snake to form the polygon that delimits the boundary of the room.

## 3.3 Transformation

This section discusses the matching process – how a subgraph is used to locate and transform a function from the case base. The matching process first finds all subgraphs in the case base isomorphic to the extracted subgraph and all possible isomorphic mappings between their nodes, using a labeling algorithm (MacDonald, 1992). The number of isomorphic mappings is potentially exponential in the number of nodes. Here, the graphs typically have only a few nodes and a few symmetries, so only a few isomorphic mappings. Associated with each node of a subgraph is an $(x, y)$ coordinate. An affine transform, Equation 2, is found that minimizes the distances between the coordinates of the mapped nodes for each of the isomorphic subgraphs. The advantage of this transform is its relative flexibility while having a simple form.

$$x' = C_0 x + C_1 y + C_2 \qquad y' = C_3 x + C_4 y + C_5 \tag{2}$$

Ideally the transformed nodes would be positioned exactly over the mapped nodes, but this is not usually possible. Even with simple rectangular shapes, the case base may not contain a graph with exactly the same doorway positions. Using a graph that is not an exact match will introduce some error in the composed function for the new task. By weighting some nodes more than others where the error occurs can be controlled. One aim is to minimize the introduction of errors that affect the overall path length. However, of equal importance is that the errors introduced be easily correctable by normal reinforcement learning.

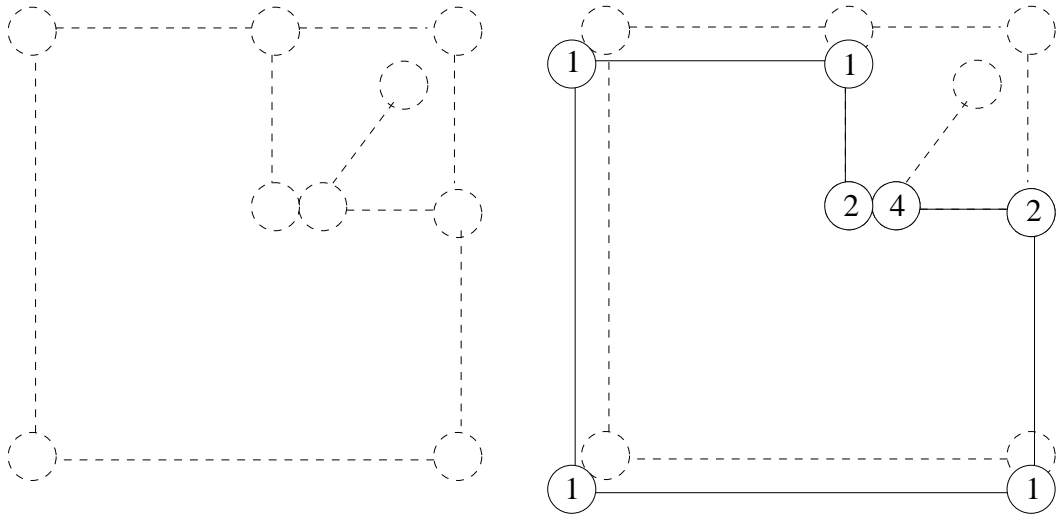

Figure 20: Weighting Graph Nodes

The left hand side of Figure 20 shows the composite graph for the new task. The right hand side shows the result of overlaying it with a graph from the case base. If the fit at the doorway of the outer L-shaped room is in error, the robot will tend to miss the doorway and collide with the wall on one side. The farther the doorway is out of position, the longer normal reinforcement learning will take to correct the error. To encourage a good fit at the doorway, a weight of 4 is used. Nodes adjacent to the doorway are given a weight of 2, all other nodes have a weight of one. This is based on the intuition that more trajectories, from different parts of the state space, will be pass through the region close to the doorway. Any error here is likely to have a broader effect, and take longer for normal reinforcement learning to correct, than in regions far from the doorway. So the fit around the inner room is improved by sacrificing fit far from the doorway.

The exact position of the doorway in the inner room is not critical and its weight is set to 0.5. Whatever the position of the doorway, the shape of the function will be correct inside the room as the goal is also in this room. However, the further the doorway is from its correct position, the greater the error in the edge length. This will produce some error in the composed function, but again the expectation is that this error will be small and reinforcement learning will quickly correct it.

Not only should the fit be good, but we would also prefer that the amount of transformation be small. All transforms produce some error and this is particularly true of asymmetric scaling, as discussed later in this section. Generally the transform produces translation, reflection, rotation, shearing and independent scaling in each dimension. In the robot navigation domain, the distance between points in the state space is just the normal Euclidean distance. The reinforcement learning function is an exponential decay in the distance to goal. If the transformation does not change the Euclidean distance, the transformed function should be directly applicable.

$$Affine \supset Similar \supset Symmetric \qquad (3)$$

The affine transformation is just one family in a hierarchy of transformations. At the bottom of this hierarchy, shown in Equation 3, are the symmetric transformations. These solid body transformations do not change the Euclidean distance. The next step up in the hierarchy introduces scaling, equal in each dimension. This will affect the Euclidean distance but only by a multiplicative factor. Thus the only change needed to the transformed function is to scale the height. The affine transformations allow the addition of asymmetric scaling and shear, which will distort the Euclidean distance. To determine the amount of distortion, the transformation is applied to the unit circle. The symmetric, rigid body, transformations will not alter the circle, but the other transformations will. The symmetric scaling transform just changes the diameter of the circle. The asymmetric scaling and shear transformations change the circle into the ellipse. The amount of distortion of the Euclidean distance introduced by the transform can be determined by the ratio of lengths of the major and minor axes of the ellipse.

$$
\begin{aligned}
error \quad = \quad & sqrt(\sum_i w_i(\Delta x_i^2 + \Delta y_i^2)) \quad && \text{(node misalignment)} \\
+ \quad & \left(\log_2 \left|\frac{r_{maj}}{r_{min}}\right|\right)^2 \quad && \text{(Euclidean Distortion)} \\
+ \quad & 0.05 \left(\log_2 \left(\frac{|r_{maj}+r_{min}|}{2}\right)\right)^2 \quad && \text{(scaling factor)}
\end{aligned}
\qquad (4)
$$

The error of fit of the transformed subgraph can be combined with the transformation error using the lengths of the major and minor axes, $r_{maj}$ and $r_{min}$ respectively, of the ellipse. There is a penalty for Euclidean Distortion from asymmetric scaling and shear. The log factor is added directly to the error of fit as shown in Equation 4. Log factors are used, so that the penalty functions are symmetric. There is a small penalty for symmetric scaling. Once the best matching subgraph has been found, the same transformation can be applied to the associated function. If no isomorphic graph is found with a total error less than 1.5, a constant function will be used as a default. Where the new graph overlays the old graph, values are assigned by using bilinear interpolation on the discrete values of the function. Where it does not, bilinear extrapolation is used, based on the closest values. In both cases once the four values are selected, the value for the new point is calculated as shown in Equation 5. As the action-value function is indexed by action as well as state, this process is carried out for each action in turn. Any rotation or reflection in the transform is also applied to a predefined matrix of actions. This produces the necessary mapping of actions from the original to the new action-value function.

$$v = c_1 x + c_2 y + c_2 xy + c_3 \tag{5}$$

Finally, the height of the new action-value function must be adjusted to account for the change in overall distance to goal. The height of the value function at an "out" doorway is $\gamma^{dg}$ where $dg$ is the distance to goal and $\gamma$ the discount factor. The value at some random point within the room is $\gamma^{dg+dd}$ where $dd$ is the distance to the doorway. The action-value function is first normalized by dividing by $\gamma^{dg}$, the height of the function at the doorway in the original problem. It is then multiplied by $\gamma^{dng}$, where $dng$ is the distance to the new goal; the value of the point becomes $\gamma^{dng+dd}$. Scaling will also affect the height of the function. Assuming the scaling is symmetric then the new value function for anywhere in the room will be $\gamma^{c*dd}$ where $c$ is the scale factor. Thus raising the function to the power of $c$ i.e. $(\gamma^{dd})^c$ will account for scaling. When scaling is symmetric the result is exact, assuming distance is based on the linear combination of the two dimensions. With asymmetric scaling, the result is not exact. But if the difference between the two scale factors is relatively small, it is a useful approximation to use their maximum.

The following is an algorithmic summary of the whole matching process:

- SG = subgraph extracted from the new task.

- For each subgraph G acting as an index to the case base

  - For each isomorphic mapping of G to SG

    * Find minimum weighted least squares fit of G to SG using mapping
    * Affine transform = coefficients of least squares fit
    * Penalized fit = least squares error + transform penalty
    * Keep graph and transform with lowest penalized fit

- Retrieve function associated with best graph from case base (if none use default)

- Apply affine transform to function

- Apply bilinear interpolation/extrapolation

- Adjust function height

- Add new function to existing function

### 3.4 Composition

This section describes function composition, how the transformation is applied successively to the series of subgraphs extracted from the composite graph. Function composition uses a slightly modified form of Dijkstra's algorithm (Dijkstra, 1959) to traverse the edges between doorway nodes. The left hand side of Figure 21 shows the composite graph after moving the goal in the robot navigation example of Section 2.2. The right hand side shows the graph traversed by Dijkstra's algorithm.

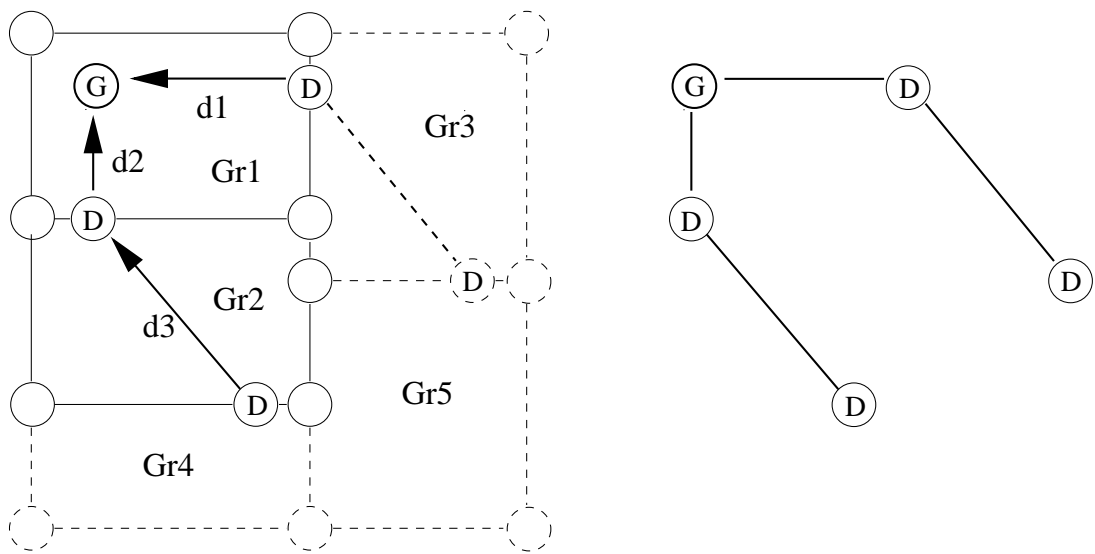

Figure 21: Using Dijkstra's Algorithm

To begin the process, the subgraph which contains the goal is extracted and the best matching isomorphic subgraph is found. The edge lengths in the composite graph are then updated using the scaled length of the corresponding edge in the matching isomorphic subgraph, d1 and d2 in Figure 21. As d2 is less than d1, the next subgraph extracted, Gr2, is the one sharing the doorway node with the edge of length d2. The best matching isomorphic subgraph is found and the edge of length d3 updated. The shortest path is again determined. As d1 is less than d2 + d3 subgraph, Gr3 is extracted. The process is repeated until all subgraphs have been updated. At each stage when a subgraph is matched, the corresponding transformed function is retrieved and added to the new function in the appropriate region.

In this example, there is only a single path to the goal from each room. Often there will be multiple paths. Suppose room 5 had an additional doorway in the lower left corner of the room, labeled "B" on the left hand side of Figure 22, in addition to the original doorway labeled "A". The graph, shown on the right hand side of Figure 22, would result. There are now two possible paths to the goal of lengths d4 and d5. If the length across room 5, d6, is greater than the absolute difference between d4 and d5, the choice of path from this room will be determined by a decision boundary inside the room. This is produced by taking the maximum of two functions as shown in Figure 23: one for entering by doorway "A" and leaving by doorway "B"; one for entering by doorway "B" and leaving by doorway "A". This principle can be repeated if there are more than two paths to the goal from a given room.

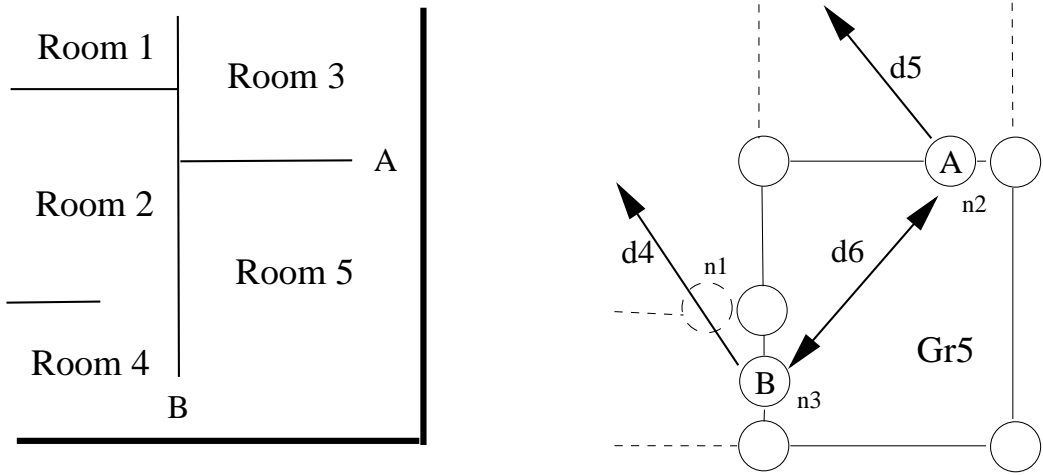

Figure 22: Multiple Paths to Goal

If the cross-room distance, d6, had been smaller than the difference (|d4-d5|) the decision boundary would have to be in another room. In general, we want to find the room in which the cross-room distance is larger than the difference between the incident paths. This is repeated for every cycle in the path graph. A cycle is detected when a node is visited twice, indicating that it is reachable by two separate paths. Let us suppose this is node n3 in the graph of Figure 22. As Dijkstra's algorithm is being used, we know that all previous nodes, on either path, such as n1 and n2 are already closed. This must be true for both paths to have reached n3. All the rooms on paths up to these nodes cannot contain the decision boundary, so it must be in either room 4 or 5. To decide which remaining room it is in, we compare the two path lengths. If d4 is longer than d5 + d6 then the decision boundary will be in room 4; otherwise it will be in room 5.

Whichever room is selected, the decision boundary is produced from the maximum of two functions. The heights of the two functions, when adjusted for their path lengths, determine where the decision boundary occurs within the room. If the paths are of equal length, taking the maximum will correctly put the decision boundary at the doorway. If there are no such functions in the case base, functions that already include decision boundaries may be used. This technique produces a correct decision boundary if the difference in the path lengths entering the room is less than the difference between the heights of the function at the "out" doorways. On the left hand side of Figure 24 there is a room with two doorways. As path 1 is significantly longer than path 2, the decision boundary is far to the left. The shortest path to the goal from most of the room is via the right hand doorway. If this function is combined with a mirror image of itself, it will produce a decision boundary in the middle

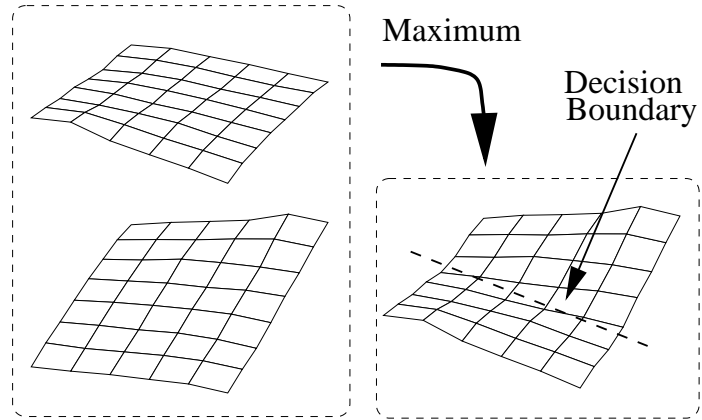

Figure 23: Combining Two Functions

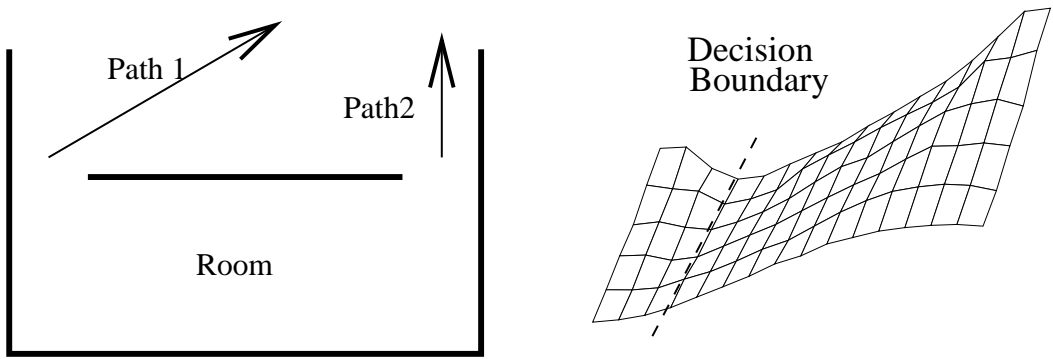

Figure 24: Decision Functions

of the room, as shown on the right hand side of Figure 25. This could be used for the new problem shown on the left hand side of Figure 25 where the two paths are the same length. Again the heights of the two functions can be changed to move the decision boundary. But it cannot be moved to anywhere in the room. The decision boundary can be moved no closer to a particular doorway than in the original function shown in Figure 24

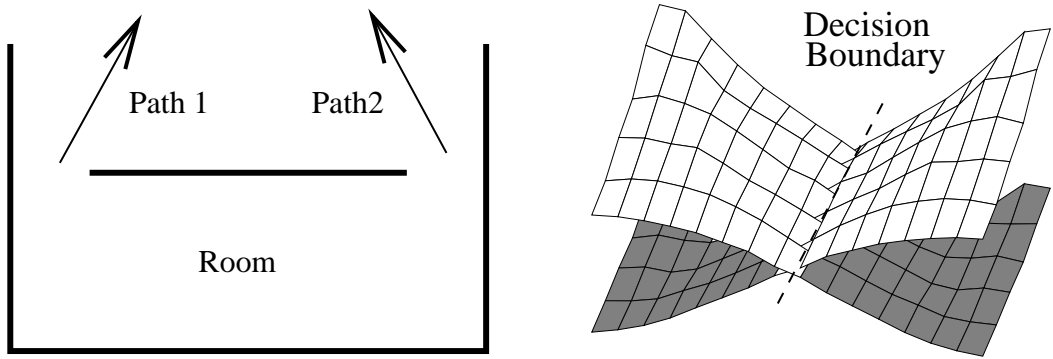

Figure 25: Combining Decision Functions

## 4. Experiments

This section compares learning curves for function composition and a simple baseline algorithm. Four sets of results are presented; one for each of the two types of related task in each of the two domains. The learning curves represent the average distance to goal as a function of the number of actions taken during learning. The distance is averaged over 64 different start positions, distributed uniformly throughout the state space, and over the different experimental runs. To determine this distance, normal learning is stopped after a fixed number of actions and a copy of the function learned so far is stored. One of the 64 start positions is selected, learning is restarted and the number of actions needed to reach the goal is recorded. If a trial takes 2000 actions and has not yet reached the goal, it is stopped and the distance to goal recorded as 2000. The function is then reinitialized with the stored version and another start state selected. This is repeated 64 times. Then the function is reinitialized once more and normal learning resumed.

The baseline algorithm and the underlying learning algorithm for the function composition system is the basic Q-learning algorithm, using a discrete function approximator as discussed in Section 3.1. The learning rate $\alpha$ is set to 0.1, the greedy policy uses an $\epsilon$ of 0.1 (the best action is selected 90% of the time), the future discount $\gamma$ is 0.8 and a reward of 1.0 is received on reaching the goal. Although the state spaces for the different domains represent two quite different things – the robot's $\langle x, y \rangle$ location and the angle of the arm's two joints – the actual representation is the same. The state space ranges between $\pm 1$ for each dimension. A step is $\pm 0.25$ in each dimension either separately or together, giving the eight possible actions. The actions are stochastic, a uniformly distributed random value between $\pm 0.125$ being added to each dimension of the action. In the robot navigation examples if the robot hits the wall, it is positioned a small distance from the wall along the direction of its last action. This has not been implemented for the robot arm as it is a somewhat more complex calculation. Instead, if a collision with an obstacle occurs the arm is restored to its position before taking the action.

Learning begins at a randomly selected start state and continues until the goal is reached. Then a new start state is selected randomly and the process repeated. This continues until the requisite total number of actions is achieved. Speed up is calculated by dividing the number of learning steps at one specific point on the baseline learning curve by the number of learning steps at an equivalent point on the function composition system's learning curve. The knee of the function composition system's curve is used. This occurs where the low level learning algorithm is initialized with the composed function. This is compared to the approximate position of the knee of the baseline curve.

## 4.1 Robot Navigation, Goal Relocation

The first experiment investigates the time taken to correct a learned function when the goal is relocated in the robot navigation domain. There are nine different room configurations, as shown in Figure 26, the number of rooms varying from three to five and there are four different goal positions. Each room has one or two doorways and one or two paths to the goal. To initialize the case base, a function is learned for each of these configurations with the goal in the position shown by the black square. The rooms are generated randomly, with some constraints on the configuration of the rooms and doorways: a room can not be too small or narrow, a doorway can not be too large. The case base also includes functions generated for the experiments discussed in Section 4.3. This was necessary to give a sufficient variety of cases to cover most of the new tasks. Even with this addition, not all subgraphs are matched. Constant valued default functions are used when there is not a match. This reduces speed up significantly, but does not eliminate it altogether.

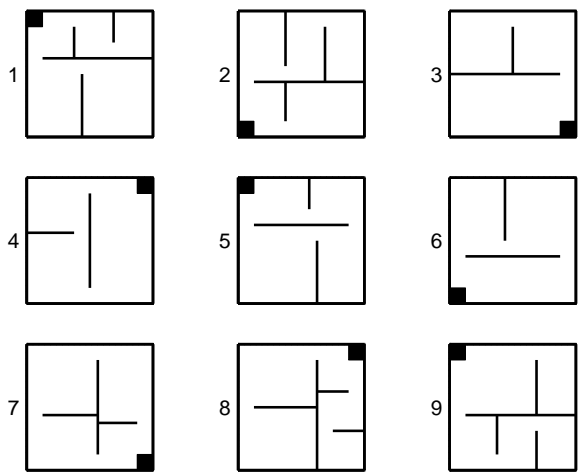

Figure 26: The Different Suites of Rooms

Once the case base is loaded, the basic Q-learning algorithm is rerun on each room configuration with the goal in the position shown. After 400,000 steps the goal is moved, this is denoted as time t on the $x$-axis of Figure 27. The goal is moved to one of the three remaining corners of the state space, a task not included in the case base. Learning continues for a further 300,000 steps. At fixed intervals, learning is stopped and the average number of steps to reach the goal is recorded. The curves in Figure 27 are the average of 27 experimental runs, three new goal positions for each of the nine room configurations.

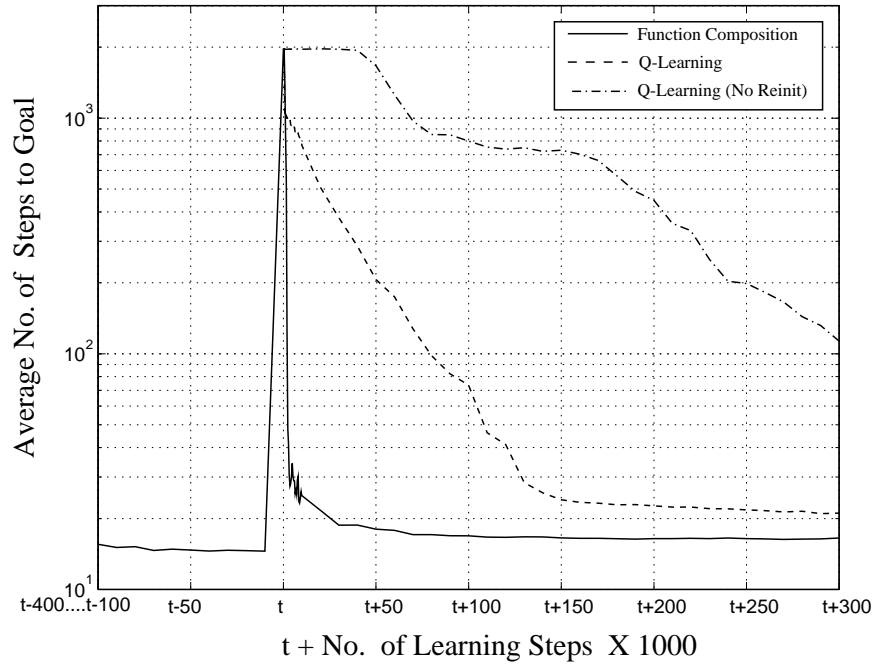

Figure 27: Learning Curves: Robot Navigation, Goal Relocation

The basic Q-learning algorithm, the top curve of Figure 27, performs poorly because, when the goal is moved, the existing function pushes the robot towards the old goal position. A variant of the basic algorithm reinitializes the function to zero everywhere on detecting that the goal has moved. This reinitialized Q-learning, the middle curve, performed much better, but it still has to learn the new task from scratch.

The function composition system, the lowest curve, performed by far the best. The precise position of the knee of this curve is difficult to determine due to the effect of using default functions. If only those examples using case base functions are considered, the knee point is very sharp at about 3000 steps. The average number of steps to goal at 3000 steps, for all examples, is 40. The non-reinitialized Q-learning fails to reach this value within 300,000 steps giving a speed of over 100. The reinitialized Q-learning reaches this value at about 120,000 steps, giving a speed up of about 40. Function composition generally produces accurate solutions. Even if some error is introduced, further Q-learning quickly refines the function towards the asymptotic value of about 17. After about 150,000 steps, normal Q-learning reaches an average value of 24 steps and then slowly refines the solution to reach an average value of 21 after 300,000 steps.

## 4.2 Robot Arm, Goal Relocation

The second experiment is essentially a repeat of the first experiment but in the robot arm domain. The initial number of steps, before the goal was moved, was reduced to 300,000 to speed up the experiments. As the arm has only two degrees of freedom, and with the restrictions discussed in Section 2.4, the number of variations is small. So only three obstacle configurations were used, constructed by hand, with two obstacles in each. To increase the number of experiments, to allow for greater statistical variation, each configuration was repeated with the goal in each of three possible positions, as shown in Figure 28. The black diamonds represent the obstacles, the black rectangles the goal. Solutions to all these tasks were loaded into the case base. When composing a function, however, the system is prevented from selecting a case that comes from the same goal and obstacle configuration.

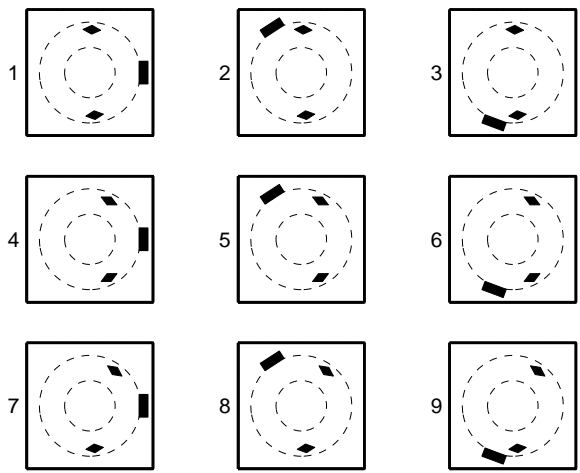

Figure 28: The Robot Arm Obstacle and Goal Positions

The curves in Figure 29 are the average of 18 experimental runs, two new goal positions for each of the three original goal positions in the three obstacle configurations shown in Figure 28. There are only two learning curves, non-reinitialized Q-Learning being dropped. As in the first experiment, the function composition system, the lower curve, performed much better than Q-learning. The knee of the function composition system occurs at 2000 steps, the knee of Q-learning at 50,000 steps, giving a speed up of 25. In this experiment, the case base contained subgraphs that matched for all new tasks, so default functions were not needed. The composed functions tend to be very accurate and little further refinement is necessary.

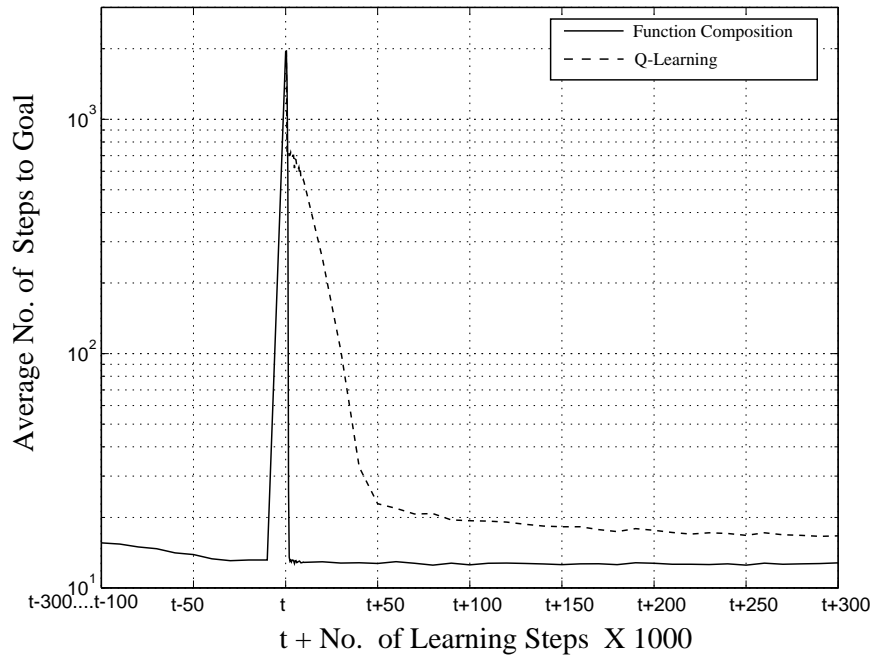

Figure 29: Learning Curves: Robot Arm, Goal Relocation

## 4.3 Robot Navigation, New Environment

The third experiment investigates the time taken to learn in a new, but related, environment in the robot navigation domain. Nine different inner rooms were generated randomly, again under some constraints. All have a single doorway, but the size and position of the room and the location of the doorway are varied as shown in Figure 30. To initialize the case base, a function is learned for each of these configurations with the goal inside the small room as indicate by the dark square. Learning is then repeated on each of the room configurations in turn. However, when composing the new function the system is prevented from selecting a case learned from the same goal and room configuration. Experimental runs for the Q-learning algorithm and the function composition system are initialized with a flat function of zero and 0.75 everywhere respectively, denoted as zero on the x-axis. Learning continues for 100,000 steps. To improve the statistical variation, experiments for each configuration were repeated three times, each time with a new random seed. The curves in Figure 31 are, therefore, the average across 27 experimental runs.

The top curve is the Q-learning algorithm, the bottom curve the function composition system. For these experiments, locating the goal took typically between 400 and 1200 steps, although some took 2000 steps. The function composition system then introduces the "no walls" function and typically a further 800 to 4000 steps are taken before usable features are generated. Again, certain experimental runs took longer, this will be discussed in Section 5.2. Due to these runs, the knee of the function composition system's curve occurs at 12,000 steps. The knee of the basic Q-learning curve occurs at approximately 54,000 steps giving a speed up of 4.5. As in previous experiments once initialized the function is very accurate and little further refinement is necessary. Basic Q-learning, on reaching the knee, takes a long time to remove the residual error.

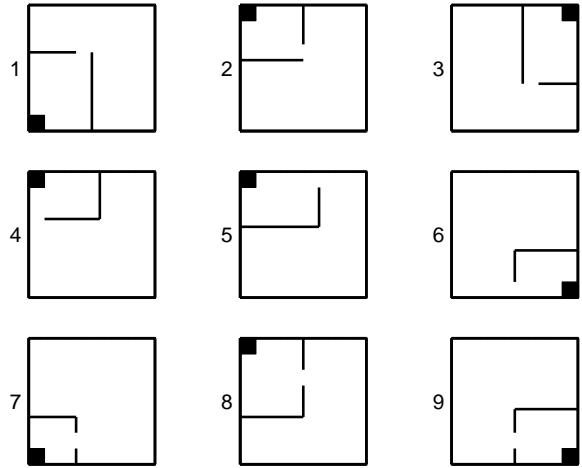

Figure 30: The Single Rooms

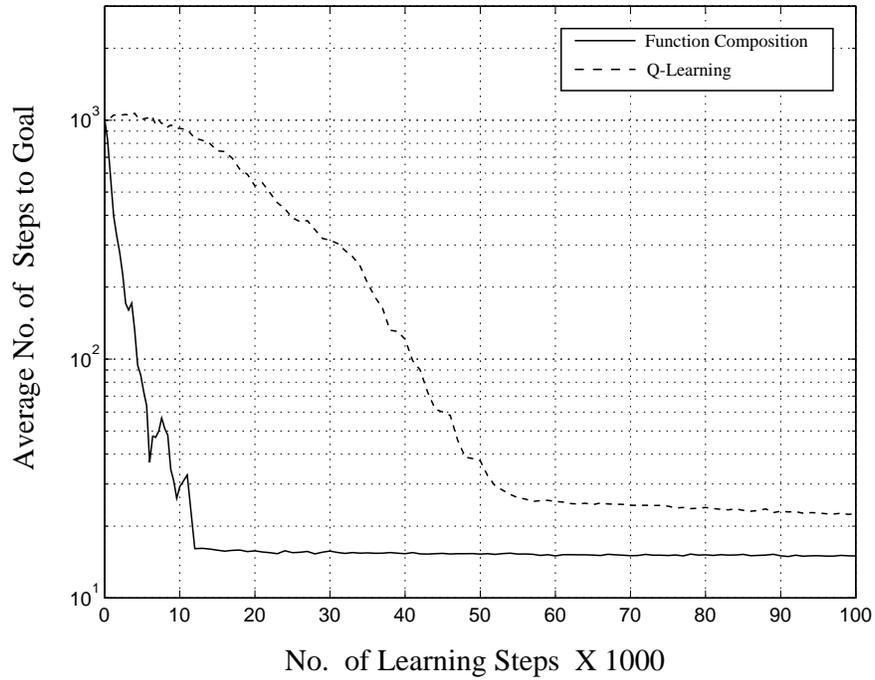

Figure 31: Learning Curves: Robot Navigation, New Environment

## 4.4 Robot Arm, New Environment

The fourth experiment is essentially the same as the third experiment except in the robot arm domain. Here, three, hand crafted, configurations of a single obstacle with the goal in a fixed position were used, as shown in Figure 32. To increase the statistical variation each configuration was run five times with a different random seed. The curves in Figure 33 are therefore the average across 15 experimental runs.

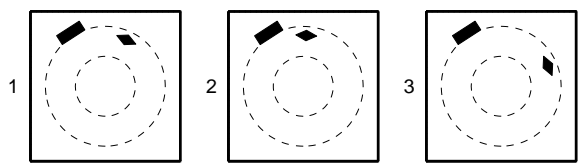

Figure 32: The Different Obstacle Positions

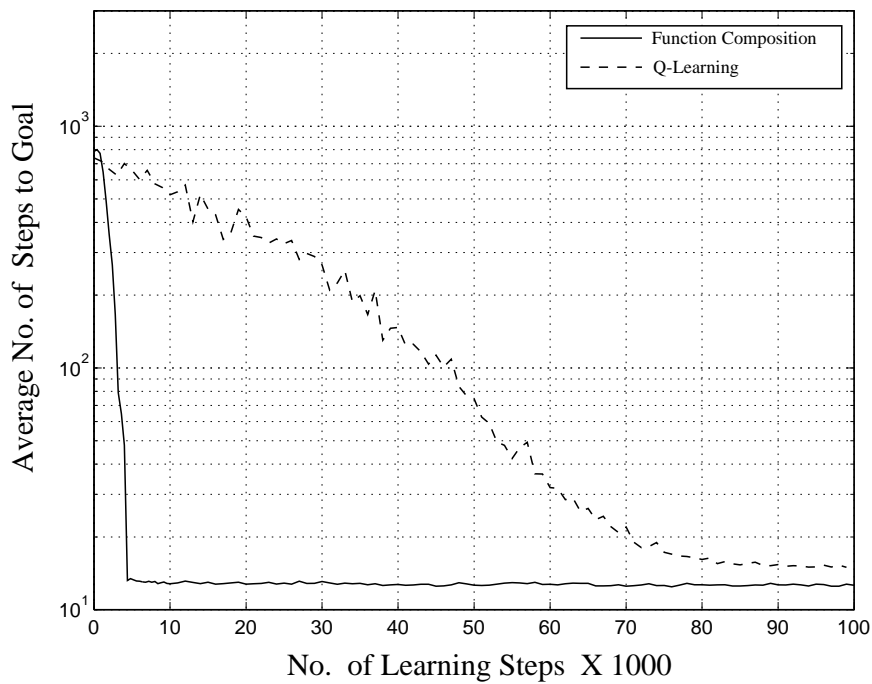

Figure 33: Learning Curves: Robot Arm, New Environment

The top curve of Figure 31 is the Q-learning algorithm, the bottom curve the function composition system. The knee of the function composition system's curve occurs at about 4400 steps. The knee of the basic Q-learning algorithm at about 68,000 steps giving a speed up of about 15.

## 5. Analysis of Results

The experiments of the previous section have shown that function composition produces a significant speed up across two different types of related task and across two domains. In addition, the composed solutions tend to be very accurate and little further refinement is required. This section begins by looking at possible concerns with the experimental methodology that might affect the measurement of speed up. It then discusses various properties of the task being solved that affect the speed up achieved by using function composition.

### 5.1 Possible Concerns with the Experimental Methodology

The speed up obtained using function composition is sufficiently large that small variations in the experimental set up should be unlikely to affect the overall result. Nevertheless, there are a number of concerns that might be raised about the experimental methodology. Some will be, at least partially, addressed in this section; others will be the subject of future work.

The first concern might be how the estimated value of speed up is measured. The value represents the speed up of the average of a set of learning tasks, rather than the average of the speed up in each of the tasks. One of the difficulties of estimation, with curves for single tasks, is that the average distance to goal may oscillate up and down as learning progresses, even though the general trend is downwards. This makes judging the position of the knee of the curves difficult, and any estimate of speed up questionable. Even experimental runs using the same configuration, but with different random seeds, exhibit a considerable variation. In some instances, the speed up measured on individual curves may benefit the function composition system, in others, the baseline algorithm. Nevertheless, probably overall most of these effects will cancel out.

The second concern might be the effect on speed up of the limit of 2000 steps when measuring the distance to the goal. Comparing two averages of values limited in this way is sometimes misleading (Gordon & Segre, 1996). But this limit primarily affects only the baseline algorithm, and was only significant when the goal was moved and the function not reinitialized. Estimation of speed up is principally concerned with comparing the position of the knees of the different curves. Here, the average distance to goal is relatively small, so limiting the value is likely to have little effect.

The third concern might be that the value of speed up is dependent on the configuration of the baseline algorithm. Certainly, it is the experience of this author that the way the function is initialized, and how actions are selected, can have an impact on the speed of learning. In previous work (Drummond, 1998), the function was initialized to a constant value of 0.75, a technique termed "optimistic initial values" by Sutton and Barto (1998). Tie breaking between actions of the same value was achieved by adding a small amount of noise (circa $\pm 5 \times 10^{-5}$). It was expected that this would increase exploration early on in the learning process and speed up learning overall. However, using an initial value of zero and a strict tie-breaker, randomly selecting amongst actions with the same value, turned out to produce a significant speed up in the baseline learning algorithm. This configuration was used for the preceding experiments, but on one experimental run this caused serious problems for the baseline algorithm.

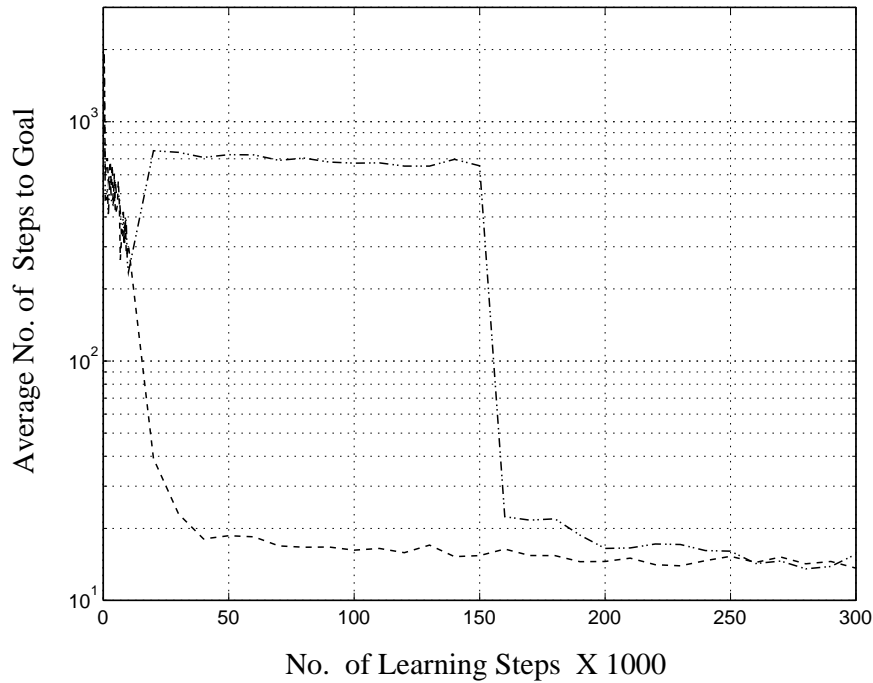

Figure 34: Learning Curves in a Partially Observable Domain

The upper learning curve of Figure 34 is for the baseline algorithm, for one run when the goal was moved in the robot arm domain. As it had such a large impact on the average learning curve, it was replaced by the lower curve, produced by repeating the experiment with a different random seed. This very slow learning rate arises from the interaction of the partial observability of the robot arm domain with the use of an initial value of zero. Individual cells of the function approximator straddle the obstacles allowing a "leak-through" of value from one side of the obstacle to the other. Starting with a zero value, once an action receives some value it will remain the best action for some time. Continual update of this action will decrease the value, but it can only asymptotically approach zero. Until other actions for the same state are updated, it will always be selected as the greedy action. This did not occur for higher initial values. It may be that in domains where there is some degree of partial observability, small initial values are better than zero or some means of improving exploration for very small values might be necessary.

Other variations in the parameters of the baseline algorithm have not been explored in this paper. For instance, a constant learning rate of 0.1 was used. Alternatives, such as starting with a higher rate and reducing it as learning progresses might also improve the overall speed of the baseline algorithm. Some preliminary experiments were, however, carried out using undiscounted reinforcement learning, the discounting being strictly unnecessary in goal-directed tasks. Room configuration 1 of Figure 26, with the goal in the lower right hand corner, was used as the experimental task. The discounting, discussed in Section 3.1, is turned off by setting $\gamma$ to 1. In addition, the value on reaching the goal state is set to zero and a cost is associated with every action. This form of learning simplifies function composition, normalization procedures needed to compensate for the value function's exponential form being no longer required. With normalization disabled, the snake successfully partitioned the function, the most critical part of the process. However, the baseline learner took considerably longer to learn the function than in the discounted case. With discounting, the learner reached an average distance to goal of about 72 steps after 80,000 learning steps. Without discounting, the learner reached an average of 400 steps at the same point in time and only an average of 80 steps after 300,000 learning steps. The action-value function was initialized to zero, which appears to be the standard practice in the literature. However, the experience with initialization in the discounted case suggests this might be the part of problem and this will be investigated in future work.

The baseline Q-learning algorithm used is the most basic and a more sophisticated one would unquestionably reduce the speed up experimentally obtained. For instance, some form of reinforcement learning using eligibility traces (Singh & Sutton, 1996) might be used. For the experiments when the goal was moved, a baseline such Dyna-Q+ (Sutton, 1990) which was specifically designed to deal with changing worlds would probably be a better reference point.

The speed up obtained, by transferring pieces of an action-value function, has also not been compared to alternatives, such as transferring pieces of a policy or transferring pieces of a model. Transferring pieces of a policy would reduce memory requirements and not require the rescaling applied to pieces of an action-value function. It does, however, have two disadvantages. Firstly, a solution can not be directly composed, as the position of decision boundaries can not be determined. Further learning would be necessary to decide the appropriate policy for each room. Secondly, the policy only indicates the best action. The action-value function orders the actions, indicating potentially useful small changes to the policy which might improve the accuracy on a new task. Transferring pieces of a model, would require first learning a model consisting of a probability distribution function for each action in each state. The memory requirement is considerably larger, unless the states reachable by an action are limited beforehand. Nevertheless, a model would need less modification in a changing world, such as when the goal is moved. It also carries more information which might speed up learning. The action-value function seems a good compromise in terms of complexity versus information content, but this would need to be empirically validated and is the subject of future work.

## 5.2 Performance Variation with Task Configuration

Generally, function composition outperforms the baseline learning algorithm by an amount dependent on the complexity of the learning problem. In the robot navigation domain when the goal was moved, the amount of speed up increased with more rooms and fewer paths to goal. A speed up of 60, against an average speed up of 40, was obtained on the configurations with five rooms and a single path to goal. Configurations with only three rooms had the least speed up, but this was not only due to the relative simplicity of the problem.

The top of Figure 35 shows the average of four learning curves for the three room configurations. The bottom of Figure 35 shows one of the configurations that produced these curves. Not only is it one of the easiest tasks (from the experimental set) for the baseline algorithm, but also there are no solutions in the case base for the lowest room. There are no isomorphic subgraphs of this form. Rather than not composing a solution, the system introduces a constant value function for this room. This room represents almost half the state space, so much additional learning is required. As the top of Figure 35 shows, initially there is significant speed up. Further refinement reduces the advantage and for a short while the baseline algorithm is better. But later, function composition gains the upper hand and converges more quickly than the baseline algorithm towards the asymptotic value.

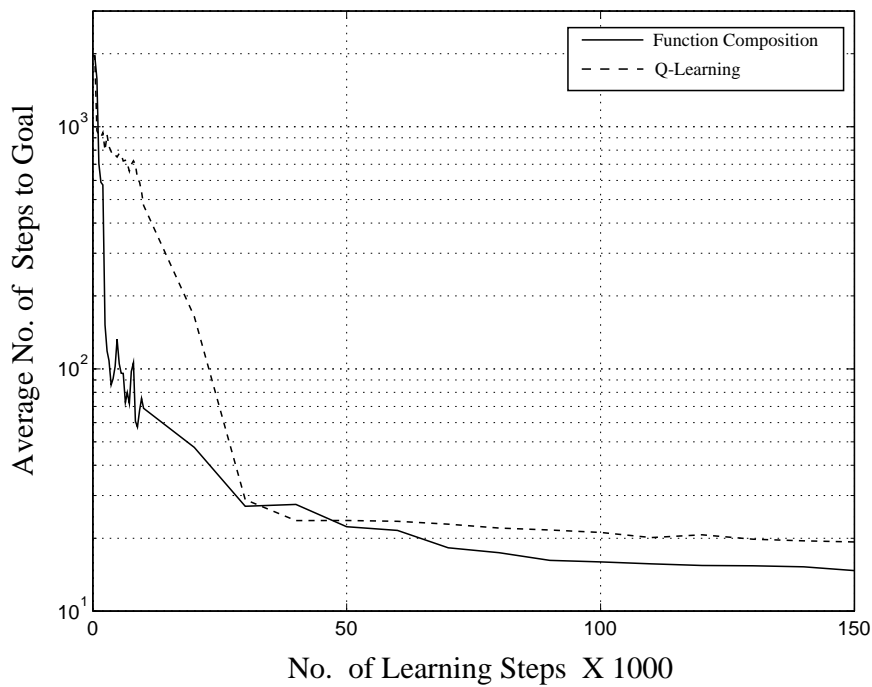

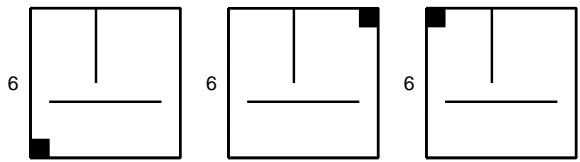

Figure 35: Failure in Robot Navigation Moving Goal

In the robot navigation domain when learning a new task, the amount of speed up varied with the size of the inner room. This was primarily due to the number of actions needed before the features emerged with sufficient clarity for the snake to locate them. Function composition is most successful when the inner room is small. If a wall is long, the feature takes more time to develop, more refinement by Q-learning is needed to make it apparent. Very short walls are also hard to identify. The likelihood of the robot colliding with them is small and it takes many exploratory actions for the features to emerge clearly.

The features may be sufficiently clear for the snake to form a partition, yet not be well enough defined to precisely locate the doorways. A doorway may appear to be a bit wider than it actually is. More importantly, it may appear to be displaced from its true position. Typically, the error in the composed function is small and normal reinforcement learning quickly eliminates it. In one of the experimental runs, configuration 2 in Figure 30, the speed up was reduced by a factor of 2 due to the doorway being incorrectly positioned. The feature representing the lower wall had not completely emerged when the partition was generated. This made the doorway appear to be almost exactly at the corner. The algorithm, in fact, positioned the doorway just on the wrong side of the corner. This resulted in the significantly reduced speed up. But it is unclear why reinforcement learning took so long to correct what seems, on the surface at least, to be a local error. This will be investigated in future work.

## 6. Limitations

Limitations come in , roughly, two kinds: those arising from the overall approach and those arising from the way it was implemented. In the former case, ways to address these limitations may be highly speculative, or impossible without abandoning some of the fundamental ideas behind the approach. In the latter case, there is a reasonable expectation that future work will address these limitations. The following sections will deal with these cases in turn.

### 6.1 Limitations in the Approach

To explore the possible limitations in the approach, this section reviews the fundamental assumptions on which it is based.

It is a fundamental assumption that features arise in the reinforcement learning function that qualitatively define its shape. The features used in this paper are the violation of a smoothness assumption, that neighboring states have very similar utility values. A wall, by preventing transitions between neighboring states, typically causes such a violation. Other things, such as actions with a significant cost, would have a similar effect. Smaller, and much more varied costs, will not generate the features required by this approach, so it offers little in the way of speed up in these cases. If there is a mixture of large and small costs, it is expected that the system will capture features generated by the former, initialize the function and normal reinforcement learning will address the latter.

The smoothness assumption is less clear if the dimensions are not numeric. The neighborhood relation, used here, is a predefined distance metric over a continuous space. In nominal, binary or mixed domains it is not obvious how such a metric would be defined, although there is some work on such metrics for other applications (Osborne & Bridge, 1997). If the dimensions are mixed, feature location might be limited to the continuous ones. If the dimensions are purely nominal or binary, a generalization of the snake may be appropriate. The snake is, at an abstract level, a constrained hill climber. But whether or not this idea would usefully generalize in this way is at present somewhat speculative.

It is a fundamental assumption that the features clearly delimit subtasks. In the domains discussed in this paper, the obstacles and walls subdivide the state space into regions connected by small "doorways". The subtask of reaching one doorway is not greatly affected by the subsequent subtask. In other domains this may not be the case. As the doorways become larger, the context sensitivity increases. As long as the composed solution is reasonably accurate, reinforcement learning can easily correct the error although speed up will be reduced. At some point however, due to a very large amount of context sensitivity, the advantage of dividing the task into subtasks will become questionable. It would be possible to account for some of the context dependency in the graph matching stage, looking at larger units than subgraphs. If two adjacent subgraphs match the new problem, they might be used as a pair, thereby including any contextual relationship between them. Even if single subgraphs were used, the context in which they appear, i.e. the shape of neighboring subgraphs, could be taken into account. In the limit, graph matching the whole task might be used. But, as was argued in the introduction, this would considerably limit when transfer is applicable, and thus its overall effectiveness.

It is a fundamental assumption that the absolute position of the features is unimportant, it is the shape of the delimited region that matters. To increase the likelihood of transfer, solutions to subtasks have been subjected to a variety of transformations. In some domains, many, if not all, of these transformations will be invalid. If actions cannot be rotated or reflected, or if many small costs affect different regions of the state space, the effectiveness of transfer will be reduced. This would be, to some extent, addressed by additional penalties for different transformations, but again this would limit the opportunities for transfer. Which transformations are appropriate, and whether or not this can be determined automatically from the domain, will be the subject of future research.

It is a fundamental assumption that a vision processing technique can locate these features in a timely fashion, even in very high dimensional domains. Learning in very high dimensional domains is likely to be slow whatever technique is used. Normal reinforcement learning will take time to navigate the much larger space, slowing down the emergence of the features. Although the time taken to partition the function will increase, the frequency with which partitioning is applicable will decrease. Thus the amortized cost will rise more slowly. Further, as high dimensional spaces are generally problematical, methods such as principal components analysis and projection pursuit (Nason, 1995) can be used to reduce dimensionality. It may prove in practice that the dimensionality which is important, and is the focus of feature extraction, is much smaller than the actual dimensionality of the space.

## 6.2 Limitations in the Implementation

If the assumptions of the previous section are met, it is expected that the remaining limitations are due to the present implementation. These limitations are likely to become apparent when the system is applied to other domains. Certainly other domains may differ from those presented in this paper in a number of ways.

A domain may differ in that the dimensionality of the space is higher than the two dimensions of the tasks investigated in this paper. The implementation of the snake has been updated to work in higher dimensions. The bold lines at the top of Figure 36 are one of the simpler tasks from the robot navigation domain. The task has been extended in the Z-dimension. The snake starts out as a sphere and then expands outwards until it fills the room. In this example, the polygonal constraint has not been used, but everything else remains the same. Figure 37 shows the complete partition of the task.

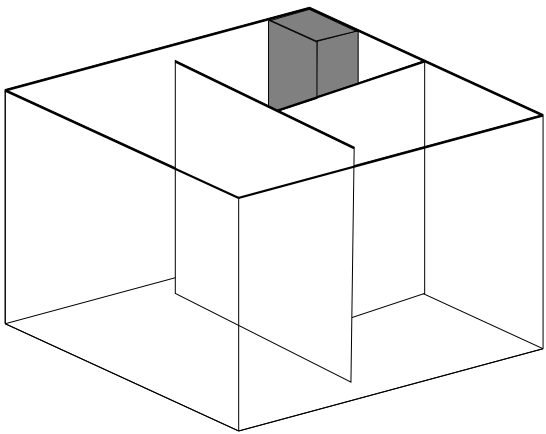
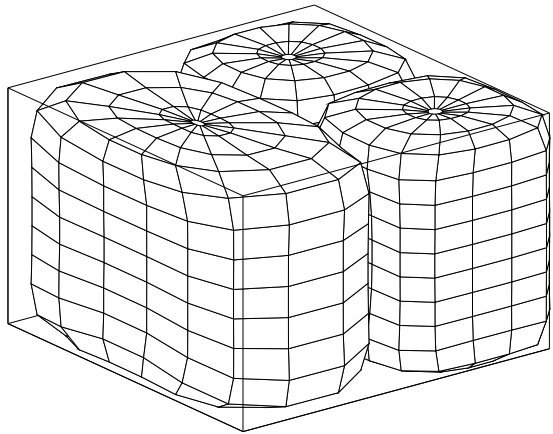

Figure 36: Adding a Z-Dimension        Figure 37: The Complete 3D Partition

The mathematics behind the snake is not limited to three dimensions. There also seems to be nothing in principle that would prevent other processes such as graph matching, planning or transformation from working in higher dimensions. Speed is the main problem. This is not a problem unique to this approach and there is a large body of research addressing this issue. For instance, although graph matching is in general NP-complete, there is much active research in speeding up matching on the average or in special cases (Gold & Rangarajan, 1996; Galil, 1986). At present, the snake represents the principal restriction on speed. This is an issue of great importance to the vision processing community. Current research is investigating this problem, at least in two or three dimensions. One example is hierarchical methods (Schnabel, 1997; Leroy, Herlin, & Cohen, 1996) which find solutions for the snake at progressively finer and finer resolution scales. The results of such research will undoubtedly be of importance here.

A domain may differ in that the value function learned might not produce features locatable by the snake with the present parameter settings. The values of the parameters were empirically determined, using hand crafted examples from the robot navigation and the robot arm domains. The obvious danger is that the parameters might be tuned to these examples. To demonstrate that this is not the case, configurations for the experiments in the robot navigation domain were generated randomly. As configurations for the robot arm domain are more tightly constrained, the hand crafted examples were used in the experiments. Nevertheless, the experiments have shown that the parameters worked successfully for random examples in the robot navigation domain. The same parameters also work successfully in the second domain, the robot arm. The following discussion demonstrates that they are also reasonably effective in a quite different domain, the "car on the hill". It is anticipated that using the results of current research into snakes will automate the selection of many parameters.

In the "car on the hill" domain (Moore, 1992), the task, simply stated, is to get a car up a steep hill, Figure 38. If the car is stationary part way up the hill, in fact anywhere within the dotted line, then it has insufficient acceleration to make it to the top. So the car must reverse down the hill and then achieve sufficient forward velocity, by accelerating down the other side, before accelerating up the hill. The state space, for the purposes of reinforcement learning, is defined by two dimensions. These are the position and velocity of the car, as shown in Figure 39. The goal is to reach the top of the hill with a small positive or negative velocity. In this domain there are two possible actions: accelerate forward, accelerate backwards. Unlike in previous domains, there is no clear mapping of the actions onto the state space. The state achieved on applying an action is determined by Newton's laws of motion. As the car has insufficient acceleration to make it up the hill from everywhere in the state space, a "wall" is effectively introduced, the bold line in Figure 39. To reach the top of the hill, the car must follow a trajectory around this "wall", the dashed line in Figure 39.

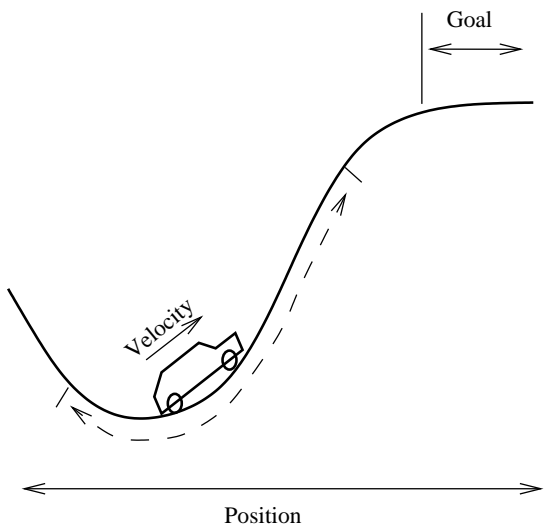

Figure 38: The Car on the Hill

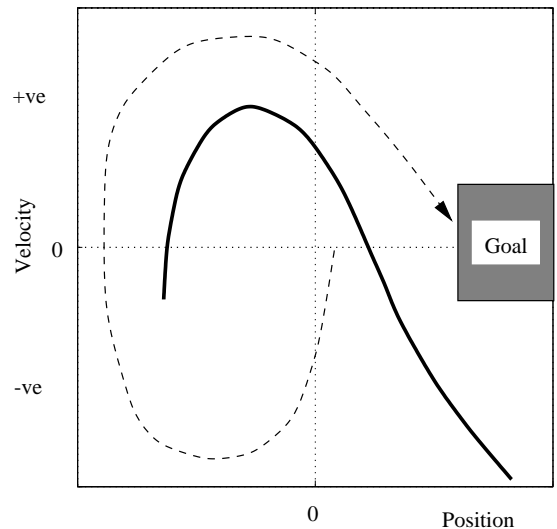

Figure 39: Car State Space

Figure 40 shows the reinforcement learning function. It exhibits the same steep gradient as the other domains. The important point to note is that, unlike in the other domains, no physical object causes this gradient. It is implicit in the problem itself, yet the features still exist. Figure 41 shows the partition produced when applying the snake to the "car on the hill" domain. The main difference from the previous examples is that the polygonal constraint has not been used. When the snake initially comes to rest, the mercury force is turned off and then the snake is allowed to find the minimum energy state. It was also necessary to reduce the scaling of the edges, by about a factor of three quarters, to achieve the accuracy of fit. The fit around the top left corner of the second snake, the dashed line, also has some problems: the snake is growing very slowly downwards and is, at present, only stopped because it has reached the maximum number of iterations allowed. One difficulty in this example is that there is not such clear delimitation of the upper and lower regions at the end of the feature. Future work will investigate altering the stopping condition to eliminate this problem.

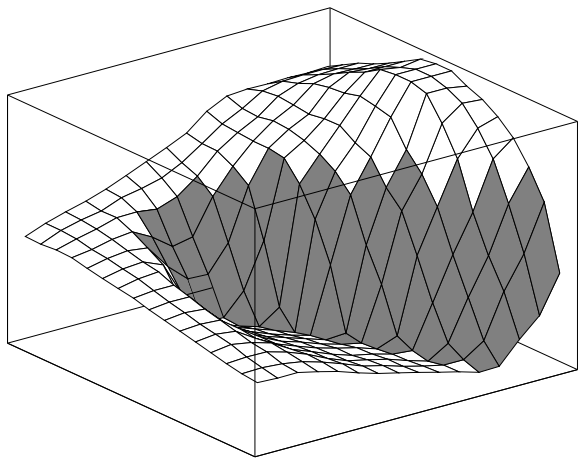

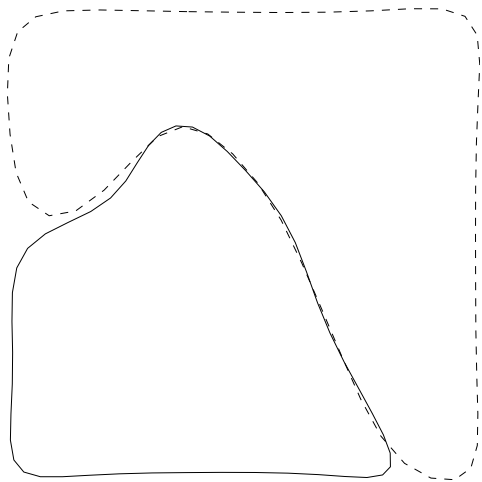

Figure 40: The Steep Gradient          Figure 41: The Regions Extracted

A domain may differ in that the shape of various regions in the partition is more complex than can be dealt with by the present snake. Fitting the snake to the task discussed in the previous paragraphs goes some way towards mitigating that concern. Nevertheless, the randomly generated examples of Section 4.1 were subject to certain constraints. Configurations with narrower rooms were tried informally, but the snake did not reliably locate the features. The configurations in Section 4 represent the limit of the complexity of partition the snake can produce at present. It is expected that using ideas from the large body of already published research into snakes will go a long way towards addressing this limitation. For complex regions, locating all the subtleties of the underlying shape may be unnecessary, or even undesirable. The aim is to speed up low level learning. As long as the solution is reasonably accurate, speed up should be obtained. Being too sensitive to minor variations in shape may severely limit the opportunities for transfer and thus reduce speed up overall.

A domain may differ in that changes in the environment are more complex than those investigated in this paper. At present, the system detects that the goal has moved by counting how often a reward is received at the old goal position. Not only is this a rather ad hoc approach, but it also does not account for other possible changes, such as paths becoming blocked or short-cuts becoming available. At present, when learning a new task the system is restarted and is not required to determine that its present solution is no longer applicable. In future work, the system should decide when its model of the world is no longer correct. It should also decide what, if any, relationship there is to the existing task and how it might be best exploited. This will allow a more complex interaction of the function composition system with reinforcement learning. For instance, the learning of a new task for the robot navigation domain used the relatively simple situation of two rooms. The function composition system initialized the low level algorithm once on detecting suitable features. In the future, to address more complex tasks, with many more rooms, an incremental approach will be used. When a new task is being learned, the system will progressively build up a solution by function composition as different features become apparent.

This approach also should handle any errors the system might make with feature extraction. In the experiments with these simple room configurations, the filtering discussed in Section 2.3 proved sufficient to prevent problems. But in more complex tasks, it is likely that false "doorways" will be detected, simply because the system has not explored that region of the state space. A composed function including that extra doorway will drive the system into that region. It should then become quickly apparent that the doorway does not exist and a new function can be composed.

## 7. Related Work

The most strongly related work is that investigating macro actions in reinforcement learning. Precup, Sutton and Singh (1997, 1998) propose a possible semantics for macro actions within the framework of normal reinforcement learning. Singh (1992) uses policies, learned to solve low level problems, as primitives for reinforcement learning at a higher level. Mahadevan and Connell (1992) use reinforcement learning in behavior based robot control. To learn a solution to a new task, all these systems require a definition for each subtask and their interrelationships in solving the compound task. The work presented here gives one way that macro actions can be extracted directly from the system's interaction with its environment, without any such hand-crafted definitions. It also shows how to determine the interrelationships of these macro actions needed to solve the new task. Thrun's research (1994) does identify macro actions, by finding commonalities in multiple tasks. But unlike the research presented here, no mapping of such actions to new tasks is proposed. Hauskrecht et al. (1998) discuss various methods of generating macro actions. Parr (1998) develops algorithms to control the caching of policies that can be used in multiple tasks. But in both cases, they need to be given a partitioning over the state space. It is the automatic generation of just such a partition that has been the focus of much of the work presented in this paper. It may well be that this approach to generating partitions and to determining the interrelationships between partitions of related tasks will prove useful to this other work.

Another group of closely connected work is the various forms of instance based or case based learning that have been used in conjunction with reinforcement learning. They have been used to address a number of issues: (1) the economical representation of the state space, (2) prioritizing states for updating and (3) dealing with hidden state. The first issue is addressed by Peng (1995) and by Tadepalli and Ok (1996) who use learned instances combined with linear regression over a set of neighboring points. Sheppard and Salzberg (1997) also use learned instances, but they are carefully selected by a genetic algorithm. The second issue is addressed by Moore and Atkeson (1993) who keep a queue of "interesting" instances, predecessors of those states where learning produces a large change in values. These are updated most frequently to improve the learning rate. The third issue is addressed by McCallum (1995b) who uses trees which expand the state representation to include prior states, removing ambiguity due to hidden states. In further work, McCallum (1995a) uses a single representation to address both the hidden state problem and the general problem of representing a large state space by using a case base of state sequences associated with various trajectories. Unlike this other research, in the work presented here the case is not an example of the value function during learning. Instead, it is the result of a complete learning episode, so the method should be complementary to these other approaches.

This work is also related to case based planning (Hammond, 1990; Veloso & Carbonell, 1993), firstly through the general connection of reinforcement learning and planning. But it is analogous in other ways. When there is a small change to the world, such as the goal being moved, a composite plan is modified by using sub-plans extracted from other composite plans.

Last, but not least, is the connection with object recognition in vision research (Suetens et al., 1992; Chin & Dyer, 1986). In the work presented here, many of the methods – if not the final application – has come from that field. The features in the reinforcement learning function are akin to edges in an image. These are located by finding the zero crossing point of the Laplacian as introduced by Marr (1982). In the work presented here, it was proposed that the features largely dictate the form of the function. Mallat and Zhong (1992) have shown that a function can be accurately reconstructed from a record of its steep slopes.

## 8. Conclusions

This paper described a system that transfers the results of prior learning to significantly speed up reinforcement learning on related tasks. Vision processing techniques are utilized to extract features from the learned function. The features are then used to index a case base and control function composition to produce a close approximation to the solution of a new task. The experiments demonstrated that function composition often produces more than an order of magnitude increase in learning rate compared to a basic reinforcement learning algorithm.

## Acknowledgements

The author would like to thank Rob Holte for many useful discussions and help in preparing this paper. This work was in part supported by scholarships from the Natural Sciences and Engineering Research Council of Canada and the Ontario Government.

## Appendix A. Spline Representations

This appendix presents some of the underlying mathematics associated with spline representations and the snake. It is not meant to be an introduction to the subject. Rather it is added for completeness to discuss certain important aspects of the system not addressed elsewhere in this paper. Knowledge of these aspects is not necessary to understand the basic principles of the approach discussed in this paper, but would be necessary if one wanted to duplicate the system. More detailed explanation is given in Drummond (1999). Some specific papers that address these ideas in much greater detail are: for splines (Terzopoulos, 1986) and for snakes (Cohen & Cohen, 1993; Leymarie & Levine, 1993).

Splines are piecewise polynomials where the degree of the polynomial determines the continuity and smoothness of the function approximation. Additional smoothing constraints can be introduced by penalty terms which reduce the size of various differentials. One way then to view spline fitting is in the form of an energy functional such as Equation 6.

$$E_{spline}(\hat{f}) = \int_R \left( E_{fit}(\hat{f}) + E_{smooth}(\hat{f}) \right) ds \tag{6}$$

Here, there is an energy associated with the goodness of fit, some measure of how close the approximating function is to the input function. This is typically the least squares distance between the functions. There is an energy associated with the smoothness of the function. Two very commonly used smoothness controls produce the membrane and thin plate splines by restricting the first and second differentials of the function respectively. To fit the spline to the function, the total energy must be minimized. A necessary condition for this is an Euler-Lagrange differential equation such as Equation 7. Here $\omega_t$ controls the tension in the spline (the resistance to stretching) and $\omega_s$ the stiffness (the resistance to bending). Often the error function will be based on individual data points and the left hand side of Equation 7 would include delta functions.

$$-\frac{\partial}{\partial s}(\omega_t(s)\frac{\partial \hat{f}(s)}{\partial s}) + \frac{\partial}{\partial s^2}(\omega_s(s)\frac{\partial^2 \hat{f}(s)}{\partial s^2}) = f_{in}(s) - \hat{f}(s) \tag{7}$$

In this work, such splines have been used for a number of purposes. When fitting the snake, measures of the first and second differential are needed. A two dimensional quadratic spline is fitted to the discrete representation of the maximum Q-values. An $\omega_t$ of 0.2 is used ($\omega_s$ is zero) to limit overshoot (Drummond, 1996) to prevent false edges. Values from an identical spline except using an $\omega_t$ of 2.0 are squared and then divided into the differential values. This normalizes the differentials, so that the size of edges is not dependent on where they occur in the function. The same type of spline is used to produce the bowls associated with the rooms as discussed in Section 3.2.1. Here $\omega_t$ is 1.0 and $\omega_s$ is 0.5 giving roughly Gaussian smoothing. The values used to produce this function are weighted. Values close to one are given weights of 200, lower values a weight of 1. This prevents the sides of the bowls from collapsing under smoothing.

A one dimensional cubic spline is used in locating the doorways. These are found by steepest descent on the value of the differential along the body of the snake. This differential contains many local minima not associated with doorways. These arise either from the inherent noise in the process or from errors of fit in the snake. The aim is to remove the ones not associated with doorways by smoothing and thresholding. This is achieved by first sampling the gradient at points along the snake. The values are then normalized to lie between zero and one. The spline has an $\omega_t$ of 0.15 ($\omega_s$ of 0.0). Here a weighted least mean squares fit is used. The weighting function is the inverse square of the values, preventing the spline from being overwhelmed by large values. Starting points for steepest descent are changes in the sign of the coefficients of the gradient of the spline. The initial step size is set to slightly larger than a knot spacing and then decreased over time. When a local minimum is found if the value exceeds a threshold (of 0.5), it is rejected.

To represent the snake, the model of the spline must be changed somewhat. The snake itself is a one dimensional cubic spline. But the energy minimum that is being sought is in the differential of the $Q_{max}$ function, subject to other constraints. The dynamics of the snake are defined by the Euler-Langrange equation shown in Equation 8.

$$\mu \frac{\partial^2 \hat{f}}{\partial t^2} + \gamma \frac{\partial \hat{f}}{\partial t} + \frac{\partial}{\partial t}\left(\frac{\partial}{\partial s^2}\left(\omega_c(s)\frac{\partial^2 \hat{f}}{\partial s^2}\right)\right) + \frac{\partial}{\partial s^2}\left(\omega_{tp}(s)\frac{\partial^2 \hat{f}}{\partial s^2}\right) = F(\hat{f}) \qquad (8)$$

An $\omega_c$ of 512 minimizes changes to the snake's shape as it grows, by penalizing the difference in the second differential to the previous time step scaled by the ratio of their lengths. An $\omega_s$ of 8.0 is the initial stiffness of the snake. This is reduced proportionately to the snake's length to give the spline more degrees of freedom. A $\mu$ of 96 and a $\gamma$ of 96 control the momentum and the drag on the snake respectively. As in Cohen and Cohen (1993), a factor is added to the energy associated with the differential that is in the direction normal to the body of the snake, as shown in Equation 9. But instead of it being a constant, a variable is used to produce the mercury model discussed in Section 3.2.1.

$$F(\hat{f}) = M(\hat{f})\,\overrightarrow{n}(s) + \nabla(-\left|\nabla Q_{max}(\hat{f})\right|^2)\,\overrightarrow{n}(s) \qquad (9)$$

The energy minimization process is carried out iteratively interleaving steps for the $x$ and $y$ directions. The differential of $\nabla |Q_{max}|^2$ for the $x$ direction is given by Equation 10, a similar equation is used for the y direction.

$$-\frac{\partial |\nabla Q_{max}|^2}{\partial x} = -2\left[(\frac{\partial Q_{max}}{\partial x})(\frac{\partial^2 Q_{max}}{\partial x^2}) + (\frac{\partial Q_{max}}{\partial y})(\frac{\partial^2 Q_{max}}{\partial x \partial y})\right] \qquad (10)$$

The snake grows under the forces of the mercury model until it reaches an approximately stable position, subject only to small oscillations. It is then converted into a polygon by finding the corners (where the normal passes through $\frac{(2n+1)\pi}{4}$ where $n = 0\ldots3$). The coefficient $\omega_1$ is set to zero everywhere. The coefficient $\omega_2$ is set to zero at the corners and 15 between them. This produces a polygon which is flexible at its vertices.

To detect the features as early as possible in the learning process, as discussed in Section 2.4, the height of the gradient is scaled according to the signal to noise ratio. The noise arises from variations in the low level learning process and the stochastic nature of the task. Both the size of the features and the noise grow with time and are somewhat normalized by this scaling process. The idea is to collect uniformly sampled values of the function shown in Equation 10 for both the $x$ and $y$ directions and find the median of their absolute values. The median is not strongly affected by extreme values and thus largely ignores the size of the features, measuring only the noise of the regions in between.

## References

Chin, C. H., & Dyer, C. R. (1986). Model-based recognition in robot vision. *Computing Surveys, 18*(1), 67–108.

Christiansen, A. D. (1992). Learning to predict in uncertain continuous tasks. In *Proceedings of the Ninth International Workshop on Machine Learning*, pp. 72–81.

Cohen, L. D., & Cohen, I. (1993). Finite element methods for active contour models and balloons for 2-d and 3-d images. *IEEE Transactions On Pattern Analysis And Machine Intelligence, 15*(11), 1131–1147.

Dijkstra, E. W. (1959). A note on two problems in connexion with graphs. *Numerische Mathematik, 1*, 269–271.

Drummond, C. (1996). Preventing overshoot of splines with application to reinforcement learning. Computer science technical report TR-96-05, School of Information Technology and Engineering, University of Ottawa, Ottawa, Ontario, Canada.

Drummond, C. (1997). Using a case-base of surfaces to speed-up reinforcement learning. In *Proceedings of the Second International Conference on Case-Based Reasoning*, Vol. 1266 of *LNAI*, pp. 435–444.

Drummond, C. (1998). Composing functions to speed up reinforcement learning in a changing world. In *Proceedings of the Tenth European Conference on Machine Learning*, Vol. 1398 of *LNAI*, pp. 370–381.

Drummond, C. (1999). *A Symbol's Role in Learning Low Level Control Functions*. Ph.D. thesis, School of Information Technology and Engineering, University of Ottawa, Ottawa, Ontario, Canada.

Galil, Z. (1986). Efficient algorithms for finding maximum matching in graphs. *ACM Computing Surveys, 18*(1), 23–38.

Gold, S., & Rangarajan, A. (1996). A graduated assignment algorithm for graph matching. *IEEE Transactions On Pattern Analysis And Machine Intelligence, 18*(4), 377–388.

Gordon, G. J. (1995). Stable function approximation in dynamic programming. In *Proceedings of the Twelfth International Conference of Machine Learning*, pp. 261–268.

Gordon, G. J., & Segre, A. M. (1996). Nonparametric statistical methods for experimental evaluations of speedup learning. In *Proceedings of the Thirteenth International Conference of Machine Learning*, pp. 200–206.

Hammond, K. J. (1990). Case-based planning: A framework for planning from experience. *The Journal of Cognitive Science, 14*(3), 385–443.

Hauskrecht, M., Meuleau, N., Boutilier, C., Kaelbling, L. P., & Dean, T. (1998). Hierarchical solution for Markov decision processes using macro-actions. In *Proceedings of the Fourteenth Conference on Uncertainty In Artificial Intelligence*, pp. 220–229.

Kass, M., Witkin, A., & Terzopoulus, D. (1987). Snakes: Active contour models. *International Journal of Computer Vision, 1*, 321–331.

Leroy, B., Herlin, I. L., & Cohen, L. D. (1996). Multi-resolution algorithms for active contour models. In *Proceedings of the Twelfth International Conference on Analysis and Optimization of Systems*, pp. 58–65.

Leymarie, F., & Levine, M. D. (1993). Tracking deformable objects in the plane using an active contour model. *IEEE Transactions On Pattern Analysis And Machine Intelligence*, *15*(6), 617–634.

MacDonald, A. (1992). Graphs: Notes on symetries, imbeddings, decompositions. Tech. rep. Electrical Engineering Department TR-92-10-AJM, Brunel University, Uxbridge, Middlesex, United Kingdom.

Mahadevan, S., & Connell, J. (1992). Automatic programming of behavior-based robots using reinforcement learning. *Artificial Intelligence*, *55*, 311–365.

Mallat, S., & Zhong, S. (1992). Characterization of signals from multiscale edges. *IEEE Transactions On Pattern Analysis And Machine Intelligence*, *14*(7), 710–732.

Marr, D. (1982). *Vision: a Computational Investigation into the Human Representation and Processing of Visual Information*. W.H. Freeman.

McCallum, R. A. (1995a). Instance-based state identification for reinforcement learning. In *Advances in Neural Information Processing Systems 7*, pp. 377–384.

McCallum, R. A. (1995b). Instance-based utile distinctions for reinforcement learning with hidden state. In *Proceedings of the Twelfth International Conference on Machine Learning*, pp. 387–395.

Moore, A. W., & Atkeson, C. G. (1993). Prioritized sweeping: Reinforcement learning with less data and less real time. *Machine Learning*, *13*, 103–130.

Moore, A. W. (1992). Variable resolution dynamic programming: Efficiently learning action maps in multivariate real-valued state spaces. In *Proceedings of the Ninth International Workshop on Machine Learning*.

Nason, G. (1995). Three-dimensional projection pursuit. Tech. rep., Department of Mathematics, University of Bristol, Bristol, United Kingdom.

Osborne, H., & Bridge, D. (1997). Similarity metrics: A formal unification of cardinal and non-cardinal similarity measures. In *Proceedings of the Second International Conference on Case-Based Reasoning*, Vol. 1266 of *LNAI*, pp. 235–244.

Parr, R. (1998). Flexible decomposition algorithms for weakly coupled Markov decision problems. In *Proceedings of the Fourteenth Conference on Uncertainty In Artificial Intelligence*, pp. 422–430.

Peng, J. (1995). Efficient memory-based dynamic programming. In *Proceedings of the Twelfth International Conference of Machine Learning*, pp. 438–439.

Precup, D., Sutton, R. S., & Singh, S. P. (1997). Planning with closed-loop macro actions. In *Working notes of the 1997 AAAI Fall Symposium on Model-directed Autonomous Systems*, pp. 70–76.

Precup, D., Sutton, R. S., & Singh, S. P. (1998). Theoretical results on reinforcement learning with temporally abstract options. In *Proceedings of the Tenth European Conference on Machine Learning*, Vol. 1398 of *LNAI*, pp. 382–393.

Schnabel, J. A. (1997). *Multi-Scale Active Shape Description in Medical Imaging*. Ph.D. thesis, University of London, London, United Kingdom.

Sheppard, J. W., & Salzberg, S. L. (1997). A teaching strategy for memory-based control. *Artificial Intelligence Review: Special Issue on Lazy Learning*, *11*, 343–370.

Singh, S. P., & Sutton, R. S. (1996). Reinforcement learning with replacing eligibility traces. *Machine Learning*, *22*, 123–158.

Singh, S. P. (1992). Reinforcement learning with a hierarchy of abstract models. In *Proceedings of the Tenth National Conference on Artificial Intelligence*, pp. 202–207.

Suetens, P., Fua, P., & Hanson, A. (1992). Computational strategies for object recognition. *Computing Surveys*, *24*(1), 5–61.

Sutton, R. S. (1990). Integrated architectures for learning, planning, and reacting based on approximating dynamic programming. In *Proceedings of the Seventh International Conference on Machine Learning*, pp. 216–224.

Sutton, R. S. (1996). Generalization in reinforcement learning: Successful examples using sparse coarse coding. In *Advances in Neural Information Processing Systems 8*, pp. 1038–1044.

Sutton, R. S., & Barto, A. G. (1998). *Reinforcement Learning: An Introduction*. MIT Press.

Tadepalli, P., & Ok, D. (1996). Scaling up average reward reinforcement learning by approximating the domain models and the value function. In *Proceedings of the Thirteenth International Conference of Machine Learning*, pp. 471–479.

Tanimoto, S. L. (1990). *The Elements of Artficial Intelligence*. W.H. Freeman.

Terzopoulos, D. (1986). Regularization of inverse visual problems involving discontinuities. *IEEE Transactions On Pattern Analysis And Machine Intelligence*, *8*(4), 413–423.

Thrun, S., & Schwartz, A. (1994). Finding structure in reinforcement learning. In *Advances in Neural Information Processing Systems 7*, pp. 385–392.

Veloso, M. M., & Carbonell, J. G. (1993). Derivational analogy in prodigy: Automating case acquisition, storage and utilization. *Machine Learning*, *10*(3), 249–278.

Watkins, C. J., & Dayan, P. (1992). Technical note: Q-learning. *Machine Learning*, *8*(3-4), 279–292.